\RequirePackage{fix-cm}
\documentclass[smallcondensed]{svjour3}

\makeatletter
\def\@author#1{\g@addto@macro\elsauthors{\normalsize%
    \def\baselinestretch{1}%
    \upshape\authorsep#1\unskip\textsuperscript{%
      \ifx\@fnmark\@empty\else\unskip\sep\@fnmark\let\sep=,\fi
      \ifx\@corref\@empty\else\unskip\sep\@corref\let\sep=,\fi
      }%
    \def\authorsep{\unskip,\space}%
    \global\let\@fnmark\@empty
    \global\let\@corref\@empty  %% Added
    \global\let\sep\@empty}%
    \@eadauthor={#1}
}
\makeatother       % onecolumn (second format)
\smartqed  % flush right qed marks, e.g. at end of proof
\usepackage{lineno,hyperref}
\usepackage{adjustbox}

\usepackage{bm}
\usepackage{mathtools}
\newcommand{\myMatrix}[1]{\bm{\mathit{#1}}}
\let\oldvec\vec
\renewcommand{\vec}[1]{\oldvec{\mathit{#1}}}
\modulolinenumbers[5]
\usepackage{enumitem}
\usepackage{afterpage}
\usepackage{longtable}
\usepackage{tabu}
\usepackage{ifthen}
\newboolean{togglecolumn}
\renewcommand{\figurename}{Fig.}

\usepackage{caption}
\DeclareCaptionLabelSeparator{tableNewline}{\par}
\usepackage{xtab}
\usepackage{pdflscape}
\usepackage{natbib}
\journalname{Artificial Intelligence Review}

\begin{document}

\makeatletter
\if@twocolumn
  \setboolean{togglecolumn}{true}
\else
  \setboolean{togglecolumn}{false}
\fi

\title{A Review of Modularization Techniques in Artificial Neural Networks%\thanks{Grants or other notes
}


\author{Mohammed Amer         \and
        Tom\'as Maul %etc.
}


\institute{Mohammed Amer \at
              School of Computer Science, University of Nottingham Malaysia Campus, Semenyih, Malaysia \\
              \email{hcxma1@nottingham.edu.my}           %  \\
           \and
           Tom\'as Maul \at
              School of Computer Science, University of Nottingham Malaysia Campus, Semenyih, Malaysia
              \email{tomas.maul@nottingham.edu.my}
}


\maketitle

\begin{abstract}
Artificial neural networks (ANNs) have achieved significant success in tackling classical and modern machine learning problems. As learning problems grow in scale and complexity, and expand into multi-disciplinary territory, a more modular approach for scaling ANNs will be needed. Modular neural networks (MNNs) are neural networks that embody the concepts and principles of modularity. MNNs adopt a large number of different techniques for achieving modularization. Previous surveys of modularization techniques are relatively scarce in their systematic analysis of MNNs, focusing mostly on empirical comparisons and lacking an extensive taxonomical framework. In this review, we aim to establish a solid taxonomy that captures the essential properties and relationships of the different variants of MNNs. Based on an investigation of the different levels at which modularization techniques act, we attempt to provide a universal and systematic framework for theorists studying MNNs, also trying along the way to emphasise the strengths and weaknesses of different modularization approaches in order to highlight good practices for neural network practitioners.
\end{abstract}

\section{Introduction}

Modularity is the property of a system whereby it can be broken down into a number of relatively independent, replicable, and composable subsystems (or modules). Although modularity usually adds overhead to system design and formation, it is often the case that a modular system is more desirable than a monolithic system that consists of one tightly coupled structure.
 
Each subsystem or module can be regarded as targeting an isolated subproblem that can be handled separately from other subproblems. This facilitates collaboration, parallelism and integrating different disciplines of expertise into the design process. As each module is concerned with a certain subtask, the modules can be designed to be loosely coupled, which enhances its fault tolerance. Also, a modular design with well defined interfaces makes it easier to scale and add more functionality without disrupting existing functions or the need for redesigning the whole system. Moreover, as modules correspond to different functions, error localization and fixing tend to be easier.
 
A modular neural network (MNN) is a neural network that embodies the concepts and practices of modularity. Essentially, an MNN can be decomposed into a number of subnetworks or modules. The criteria for this decomposition may differ from one system to another, based on the level at which modularity is applied and the perspective of decomposition.

Since the inception of artificial neural networks and throughout their development, many of their design principles, including modularity, have been adapted from biology. Biological design principles have been shaped and explored by evolution for billions of years and this contributes to their stability and robustness. Evolutionary solutions are often innovative and exhibit unexpected shortcuts or trade-offs that, even if not directly implementable, often provide useful insights. Mapping from biological principles to in-silico realizations is not a linear one-to-one process. However, there are several common steps that can be shared by different such realizations. Ideally, the process of adapting a biological design principle to an artificial neural network implementation starts with identifying the key function(s) underlying the design principle. Usually, there are several complex biological details that are irrelevant to the functional essence of the principle, which can thus be abstracted away. This may be followed by some enhancement of the identified function(s) in the artificial domain. Finally, the abstracted and enhanced principle is mapped to an artificial neural network construct using a flexible platform. Convolutional neural networks (CNNs) are a poignant success story of the adoption of some of the key design principles of the visual cortex. The model of the visual cortex was greatly simplified by CNNs by eliminating complexities like the existence of different cortical areas (e.g. areas V1 and V2) and pathways (e.g. ventral and dorsal streams) and focusing on receptive field, pattern specific regions and a hierarchy of extracted features. These were realized using linear filters, weight sharing between neurons and deep composition of layers. More examples include recurrent neural networks (RNNs), which are inspired by the brain's recurrent circuits \citep{Douglas2007}, and parallel circuit (PC) neural networks \citep{KienTuongPhan2015}, which take their inspiration from retinal microcircuits \citep{Gollisch2010}.

Biological nervous systems, the early inspiration behind neural networks, exhibit highly modular structure at different levels, from synapses \citep{Kastellakis2015}, to cortical columnar structures \citep{Mountcastle1997}, to anatomical \citep{Chen2008} and functional \citep{Schwarz2008} areas at the macro level . It has been proposed that natural selection for evolvability would promote modular structure formation \citep{Clune2013}. Modularity is evolvable as it allows for further evolutionary changes without disrupting the existing functionality. It also facilitates exaptation, where existing structures and mechanisms can be reassigned to new tasks \citep{Kashtan2005}. Recognition of this prevalence of modularity in biological neural systems has left an indelible albeit somewhat irregular mark on the history of artificial neural networks (ANNs), mostly under the guise of biologically plausible models. However, with the divergence between the fields of ANNs and Neuroscience, the ANN approach has tended to become more engineering oriented, getting most of its inspiration and breakthroughs from statistics and optimization. 

Many researchers have long been aware of the importance and necessity of modularity. For example, the ANN community has for many years recognized the importance of constraining neural network architectures in order to decrease system entropy, lower the number of free parameters to be optimized by learning, and consequently, have good generalization performance. In \citet{Happel1994}, it is argued that constraining neural architectures through modularity, facilitates learning by excluding undesirable input/output mappings, by using prior knowledge to narrow the learning search space. In \citet{Caelli1999}, modularity is considered a crucial design principle if neural networks are to be applied to large scale problems. In \citet{Sharkey1996,Xu1992}, some of the early techniques of integrating different architectures or modules to build MNNs are discussed. In \citet{Caelli1999}, six different MNN models were analytically dissected, in an attempt to provide several modeling practices. On the other hand, in another comparative study \citep{Auda1998}, ten different MNN models were empirically compared using two different datasets. In \citet{Auda1999}, a survey was done about MNNs, where the MNN design process was broken down into three stages, starting by task decomposition, then training and finally, decision making.

Despite this early interest in neural network modularity, previous research has generally focused on particular MNN models and has lacked systematic principles and a broad general perspective on the topic. Previous research has also been lacking in terms of a systematic analysis of the advantages and disadvantages of different approaches, with an increased focus on empirical comparisons of very specific models. Even for theoretically focused reviews, the taxonomy is sparse and fails to capture important properties and abstractions. Moreover, the scope of modularity focused on is very narrow, ignoring important forms of modularity and focusing mainly on ensembles and simple combinations of models. These limitations need to be addressed if modularity is to be applied more generally. More general insights and a toolbox of modularity-related techniques are needed for consistently implementing successful MNNs. Fortunately, recent MNN techniques have been devised and revisited, specially in the last decade after the revival of the ANN field in the form of deep learning. 

In this review, we aim to expand previous reviews by introducing and analysing modularization techniques in the neural networks literature in an attempt to provide best practices to harness the advantages of modular neural networks. We reviewed prominent modular neural networks throughout the literature, inspected the different levels at which modularity is implemented and how this affects neural network behaviour. We then systematically grouped these techniques according to the aspect of neural networks they exploit in order to achieve modularity. Unlike previous reviews, our focus is the general systematic principles that governs applying modularity to artificial neural networks and the advantages and disadvantages of the different techniques. We produced a general taxonomy that captures the major traits of different modular neural networks at different levels and for various modularity forms and a framework that captures the essentials of the process of building a modular neural network.
 
From our study of modular neural networks in the literature, we classified modularization techniques into four major classes, where each class represents the neural network attribute manipulated by the technique to achieve modularity. We thus categorized MNN operations into the following four classes:

\begin{enumerate}
\item Domain: this is the input space or the data an MNN operates on, which in turn defines and constrains the problem we are trying to address.
\item Topology: this corresponds to an MNN's architecture, which reflects the family of models that an MNN expresses.
\item Formation: this is how an MNN is constructed and what process is used to form its components.
\item Integration: this is how the different components of an MNN are composed and glued together to form a full network.
\end{enumerate}

So, modularization techniques operating on the domain tend to act by finding a good partitioning of the input data, to which different modules can be assigned. This is the only modular level that is optional in the sense that you may have an MNN that doesn’t have an explicit modularization of the domain, however, any neural network that is modular must use at least one technique from each successive level, which includes selecting a certain modular topology, a formation technique for building the modular architecture, and an integration scheme for combining the different modules. So, as mentioned, topological modularization is the next level at which modularity is achieved, where the technique is essentially a specification of modular topology. Every topological technique is a blueprint for the structure of the MNN, and therefore defines how nodes and modules are connected. Although the topological technique specifies how the MNN as a whole should be at the end, it doesn’t specify how this architecture can be built. This is what formational techniques try to address. Formational techniques are the processes by which modular topologies can be constructed. Finally, while formational techniques focus on the building of modularity, integration techniques specify how different modules can be integrated together to achieve the desired system outputs. So, every modular neural network realization can be seen as chain of modularization techniques applied to each level or aspect of the network.

\section{Modularity}

In the domain of neural networks, modularity is the property of a network that makes it decomposable into multiple subnetworks based on connectivity patterns. It can be argued that the shift of thinking towards functional modularity in the brain and biological neural networks, is one of the greatest leaps in Neuroscience since the neuron doctrine \citep{Lopez-Munoz2006}. The concept of emphasising the importance of relative connections between neurons and that functionality emerges from intra-modular and inter-modular interactions revolutionized the way we research nervous systems and transformed the idea of a connectome \citep{Bullmore2011,Sporns2011} into a key area of brain research.
 
As already mentioned, the brain has been shown to be modular at different spatial scales, from the micro level of synapses to the macro level of brain regions. At the level of synapses, it has been suggested \citep{Kastellakis2015} that synapses show both anatomical and functional clustering on dendritic branches, and this plays a central role in memory formation. At a larger spatial scale, cortical minicolumns \citep{Buxhoeveden2002} have been suggested to be the basic building unit of the cortex, largely supported by the claim that they have all of the elements of the cortex represented within them. Lesion studies, brain imaging using fMRI and several other techniques have shown strong evidence of brain modularity, where different areas and regions of the  brain are specialized into certain cognitive or physiological functions. More recently, the pioneering work by Sporns, Bullmore and others and the introduction of graph theory into the study of brain networks have shed light on the small world nature of brain connectivity \citep{Bullmore2009,Sporns2004}. A small world network is a network characterised mainly by clusters which are groups of neurons having more interconnections within the cluster than would be expected by chance, while there is still sparse connectivity between different such clusters. Moreover, despite large networks and sparse inter-cluster connectivity, there is still a short average path length between neurons. In \citet{Sporns2004}, it was observed that there is a direct correlation between clustering and path length, and between these two measures and brain area functionality. It was suggested that areas with short average path length and low clustering tend to be integrative multimodal association areas, while those with long average path length and high clustering tend to be specialised unimodal processing areas. 

The graph theoretical approach to studying neural networks considers the network as a connected graph, where the neurons are represented by nodes (or vertices) and the synaptic connections between neurons as edges. Although, in practise, neural networks are directed graphs, i.e. edges have directionality from pre to postsynaptic neurons, for simplicity and tractability, most researchers in this area treat neural networks as undirected graphs. Central to quantifying the small world properties of biological neural networks is how to cluster or partition the nodes into modules, where each module has dense connectivity between its nodes and sparse connectivity with nodes in other modules. There is no single most efficient algorithm for solving this problem, and indeed it was proven to be an NP-complete problem \citep{Brandes2008}, however, similar problems have long been studied in Computer science and Sociology \citep{Newman2004,Newman2006}.

In the field of computer science, graph partitioning is a well studied problem, where given a certain graph and pre-specified number of groups, the problem is to equally partition the vertices into the specified number of groups, whilst minimizing the number of edges between groups. The problem was motivated by other applications before the interest in partitioning neural networks, like partitioning tasks between parallel processors whilst minimizing inter-processor communication. The main approach in computer science is a collection of algorithms known as iterative bisection, such as spectral bisection and Kernighan-Lin algorithm. In iterative bisection, first the graph is partitioned into the best two groups, then subdivisions are iteratively made until the desired number of groups is reached. The problem with these methods is that the number and sizes of groups are not known a priori when partitioning neural networks. Moreover, a lack of good partitioning measures  leads these algorithms to deterministically partition the graph into the desired number of groups, even if the partitions don’t reflect the real structure of the graph.
 
On the other hand, sociological approaches have focused more on the problem of community structure detection, which is more suited to neural network research. Community structure detection consists of the analysis of a network in an attempt to detect communities or modules, where the algorithm does not pre-specify the number or size of groups. In other words, it is an exploratory approach, where the algorithm may detect subgraphs or may signal that the graph is not decomposable. The main technique used so far in sociological studies is hierarchical clustering. Based on a metric called similarity measure, hierarchical clustering constructs a tree-like structure of network components called a dendrogram. The horizontal section of this dendrogram at any level gives the network components produced by the algorithm. The algorithm doesn’t require pre-specification of the number or sizes of groups, but it doesn’t necessarily guarantee the best division.

\begin{figure*}
\centering

\includegraphics[width=.7\textwidth]{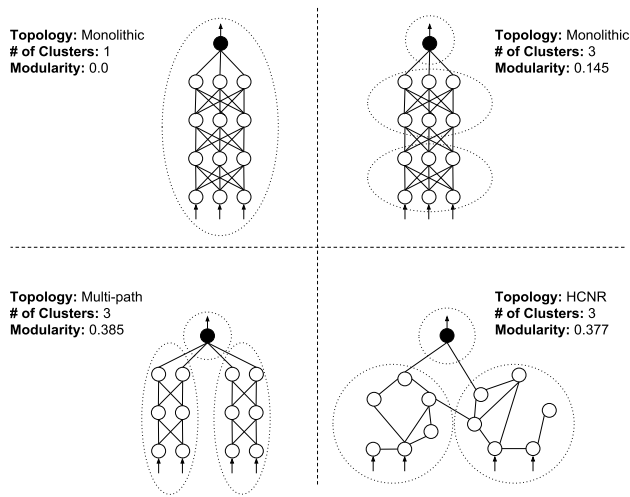}
\caption{Calculated modularity measure \citep{Newman2004,Newman2016} for different architectures. The partitions, marked as circles, used for calculations may not be optimal.}
\label{fig:mod-measure}

\end{figure*}

In more recent approaches \citep{Newman2004,Tyler2005,Radicchi2003}, a modularity measure (\autoref{eq:q}) was used to guide the detection process towards the best division. The intuitive notion of modularity as defined by \citet{Newman2004,Newman2016} is that a good network division is one that places most of the network edges within groups, whilst minimizing the number of edges between groups. Network connectivity is assumed to be described by a real symmetric matrix called adjacency matrix $\myMatrix{A}$, with dimensions $n \times n$, where $n$ is the number of nodes in the network. Each element $\myMatrix{A}_{ij}$ is $1$ if there is an edge between node $i$ and $j$ and $0$ otherwise. If we assume dividing the network into $q$ number of groups, where $g_i$ refers to the group to which node $i$ was assigned, then the sum of edges within groups (i.e between nodes of the same group) is $\frac{1}{2}\sum_{ij}\myMatrix{A}_{ij}\delta_{g_ig_j}$, where $\delta_{g_ig_j}$ is the Kronecker delta. Maximizing this quantity alone is no guide towards a good division, because assigning all the nodes to one big group would maximize this measure whilst completely avoiding any partitions. To remedy this, modularity is taken to be the difference between this quantity (i.e the actual sum of edges within groups) and the expected number of this sum if edges were placed randomly, whilst keeping the same partition. If the probability of node $i$ connecting to node $j$ after randomization is $P_{ij}$, then this expected sum is $\frac{1}{2}\sum_{ij}P_{ij}\delta_{g_ig_j}$, and the modularity measure is then

\begin{equation} \label{eq:q}
Q = \frac{1}{2m}\sum_{ij}(\myMatrix{A}_{ij} - P_{ij})\delta_{g_ig_j}
\end{equation}

where $m$ is the total number of edges, which is used as a normalization factor. The most used randomization scheme is one that preserves node degree (i.e number of edges attached to each node), and the probability of connecting node $i$ and $j$ under this scheme is $\frac{k_ik_j}{2m}$, where $k_i$ is the degree of node $i$. Please refer to \autoref{fig:mod-measure} for an illustration of different neural topologies with corresponding modularity measures. 

On the evolutionary side, multiple hypotheses have been proposed for explaining the origin of modularity in brain organization. The issue is important from the ANN perspective, as it provides inspiration for guiding evolutionary computational algorithms towards generating modular architectures. It was suggested that evolution in an environment with Modularly Varying Goals (MVGs) leads to modular networks \citep{Kashtan2005}. The MVG environment consists of varying goals with common subgoals. As the modularity of the solution obtained was usually limited, it was argued that the failure might be explained by the fact that the evolutionary algorithm was directed more towards optimal solutions, which was sufficient to solve simple problems, but due to lack of evolvability, failed to scale up to more complex problems. In other studies, the competition between efficient information transfer and wiring cost of the brain have been suggested as sufficient evolutionary pressures for modularity \citep{Clune2013,Bullmore2009}. It was also suggested that selection for minimal connection cost bootstrapped modularity, while MVG helped in maintaining it \citep{Clune2013}.

Using multiple modules in practice is partly motivated by the existence of different subproblems, that may have different characteristics, promoting problem decomposition and functional separation of tasks, that typically contribute towards maintainability and ease of debugging. There are, however, difficulties that surround applying modular neural networks to practical problems. First of all, domain decomposition into meaningful subproblems is usually difficult as the problems tackled by neural networks are usually poorly understood. Moreover, adding modularity to a neural network tends to add a number of new hyperparameters that need optimizing, such as the number of modules, the size of each module, and the pattern of connectivity between modules. Another problem that arises with multiple modules is how to integrate the output of different modules and how to resolve any decision conflicts that might arise. We try to address these different problems throughout this study. We discuss the issues surrounding domain decomposition in \autoref{sec:domain}, where we show that problem decomposition can be done implicitly or explicitly, and indicate how the process can be automated. Hyperparameter selection and associated issues are discussed in \autoref{sec:form}, where the different techniques for MNNs formation are presented. Integrating different modules to solve the task at hand is further investigated in \autoref{sec:integ}.

While the study of modularity has focused mainly on topology, which is indeed the main property of modular structure, we expand our study of modularization techniques to different levels of neural network design that can be exploited to produce modular networks. In the following section, we discuss the different levels of modular neural networks, and how different chains of techniques applied to such levels can produce different MNN variants.

\section{Modularization Techniques}

\begin{figure}[ht]
\centering

\includegraphics[width=.4\textwidth]{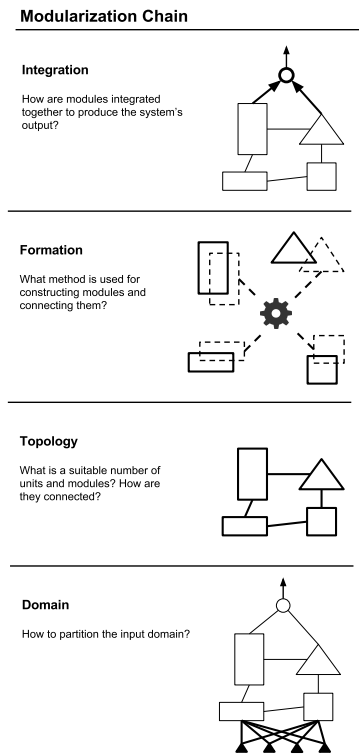}
\caption{Modularization chain acting on the different levels of a neural network.}
\label{fig:mod-chain}

\end{figure}

The modularization of neural networks can be realized with different techniques, acting at different levels of abstraction. A modularization technique is a technique applied to one of these levels in order to introduce modularity to the network topology. We present a taxonomy of such techniques that are categorized based on the abstraction level they exploit to achieve modularity. We analyse each technique, explaining the main rationale behind it, presenting its advantages and disadvantages (\autoref{tab:adv-disadv}) relative to other techniques and providing prominent use cases from the literature. The main levels at which the modularization techniques act are complementary. Consequently, to produce a modular neural network, a chain of techniques, or chain of modularization, is used. A modularization chain (\autoref{fig:mod-chain}) consists of a set of techniques (each one corresponding to a different level of the neural network environment) used to produce a modular neural network. So, every modularization chain corresponds to a particular type of MNN.

A modularization chain starts with partitioning the domain, however this is optional as was mentioned earlier. Then, a modular topological structure is selected for the model. After that, formation and integration techniques are selected to build the model and integrate the different modules, respectively. So, for example, if we need to develop an MNN for enhanced MNIST classification, then a modularization chain would look like the following:

\begin{enumerate}
\item Domain: we may choose to augment the MNIST dataset by applying a certain image processing function to a copy of each image to extract specific information, and then consider the original and processed images as different subdomains.
\item Topology: here we may select a multi-path topology, where one path of the network has the original image as input and the others have the processed ones.
\item Formation: we may use an evolutionary algorithm to build the multi-path topology, constraining it to have exactly two paths.
\item Integration: here we may integrate the outputs of each path into the final system output, either through the evolutionary process itself, or as a post-formation learning (or fine-tuning) algorithm.
\end{enumerate}

The underlying concept in the example above is to use MNNs to integrate different sources of information (i.e source and processed images) to improve classification performance. A similar concrete MNN application was investigated in \citet{Ciregan2012,Wang2015}.

\subsection{Domain} \label{sec:domain}

The domain refers to all of the information that is relevant to the problem and is accessible to the neural network learning model. In other words, it consists of the inputs and outputs that the system relies on for learning how to generalize to unseen inputs. Focusing on the input side, one of the rationales behind domain modularization, is that some functions can be defined piecewise, with each subfunction acting on a different domain subspace. So, instead of learning or applying the neural network model on all of the input space, domain modularization aims to partition this space into a set of subspaces. Then, the modules of an MNN, constructed by applying techniques at different modularization levels, can be readily applied to each subdomain. So, for example, we may choose to partition temporal data according to the time intervals in which they were collected, or partition spatial data according to the places in which they occurred like in \citet{Vlahogianni2007}. We refer to this kind of domain partitioning as subspatial domain partitioning, because the individual data items are clustered into multiple subspaces. Curriculum learning \citep{Bengio2009,Bengio2015} is a particular form of this partitioning, where the neural network is successively trained on a sequence of subspaces with increasing complexity. Another kind is what we call feature or dimensional domain partitioning. In feature domain partitioning, partitioning occurs at the level of a data instance, such that different subsets of features or dimensions or transformations of these get assigned to different partitions. Examples of this approach include the application of different filters to the original images and processing each with different modules like in \citet{Ciregan2012}, and the autocropping of image parts to augment the data and thus improve generalization \citep{Zhang2014}.

The domain is, conceptually, the most natural and straightforward level of modularization. This is essentially because the domain defines the problem and its constraints, so, a good modularization of the domain corresponds directly to good problem decomposition. Decomposition of complex problems greatly simplifies reaching solutions, facilitates the design and makes it more parallelizable in both conception and implementation. In \citet{DeNardi}, the problem of replacing a manually designed helicopter control system by a neural network couldn't be tackled when a single MLP was trained to replace the whole system. However, it was feasible by replacing the system components gradually. Moreover, as it holds all the available information about the problem and its structure, it acts as a very good hook for integrating prior knowledge, through the implementation of modularity, that may be useful in facilitating problem solving. Prior knowledge at the domain level is mainly a basis for problem decomposition, be it an analytical solution, some heuristic or even a learning algorithm. For example, in \citet{Babaei2010}, the problem domain of predicting protein secondary structure was decomposed into two main groups of factors, namely: strong correlations between adjacent secondary structure elements and distant interactions between amino acids. Two different recurrent neural networks (RNNs) were used to model each group of factors before integrating both to produce the final prediction. It is also interesting to note that the domain is the only level of modularization that can be absent, at least explicitly, from a modular neural network. In other words, you can have a modular neural network that doesn’t involve any explicit modularization at the domain level, but this is not possible for the other levels. This is mainly because domain decomposition is a kind of priming technique for modularity, in other words, it promotes modularity but is not a necessary condition. Note that domain decomposition can still happen implicitly without intentional intervention. As a simple example, it is well established in the machine learning literature that radial basis function (RBF) networks, and also non-linearities in feedforward networks, are able to transform inputs that may be non-linearly separable into linearly separable ones.

\subsubsection{Manual}

Manual domain modularization is usually done by partitioning the data into either overlapping or disjoint subspaces, based on some heuristic, expert knowledge or analytical solution. Theses partitions are then translated into a full modular solution via different approaches throughout the modularization process.
 
The manual partitioning of input space allows for the integration of prior knowledge for problems that are easily decomposable based on some rationale. This knowledge-based integration can be done by defining partitions that correspond to simple subproblems that can be addressed separately. Moreover, it gives fine control over the partitioning process that can be exploited to enhance performance. This is contrary to automatic decomposition, which may be adaptive and efficient, but as the rationale is latent and not directly observed, it is hard to tweak manually for further enhancements. On the other hand, it raises the question of what defines a good partition, a partition which corresponds to a well defined subproblem, that has isolated constraints and can be solved separately. Although the domain is what characterises a problem's solution, usually the relation between decomposing the domain and obtaining a solution is not that straightforward. For example, you may think of decomposing some input image to facilitate face recognition. However, since the process of face recognition is not well understood, it is not clear what decomposition is suitable. Is it segmentation of face parts or maybe some filter transformation? \citep{Chihaoui2016} Figuring this out analytically is not feasible. The data generating process is often very complex and contains many latent factors of variations, which makes the separation and identification of those factors hard. A good partitioning requires a good prior understanding of the problem and its constraints, which is rarely the case for machine learning tasks, which generally rely on large datasets for the automatic extraction of the underlying causal factors of the data.
 
One of the simplest subspatial partitioning schemas that arises naturally in classification tasks is class partitioning. Class partitioning is the partitioning of the domain based on the target classes of the problem. This is a straightforward approach which is built on the assumption that different classes define good partitions. There are three main class partitioning schemes, namely OAA, OAO and PAQ \citep{Ou2007}. In the One-Against-All (OAA) \citep{Anand1995,Oh2002} scheme, the domain of $K$ classes is partitioned into $K$ subproblems, where each subproblem is concerned with how to differentiate a particular class A from its complement, that is, all of the remaining classes which are not A. One-Against-One (OAO) \citep{Rudasi1991} partitions the domain into $\binom{K}{2}$ subproblems, with each subproblem concerned with differentiating one class from only one other class. A compromise between the two previous schemes is P-Against-Q (PAQ) \citep{Subirats2010}, where each subproblem aims to differentiate $P$ number of classes from $Q$ number of classes. OAO is the most divisive of the three, which makes it the most computationally expensive, assuming that each classifier's complexity is the same. Whatever the scheme used, the output of each module trained on a different subproblem can then be combined with an integration technique to embody a particular MNN. More generally, different modularization chains can be applied to the different subproblems, thus resulting in different MNNs.

Class partitioning reduces classification complexity and can be seen as a divide-and-conquer approach. If the partitioning results in several smaller datasets, and assuming that the partitioning accurately reflects the problem's underlying structure, then not only should the learning problem be easier, but the overall representation learnt by the MNN should be more faithful to the underlying causes of the data. In \citet{Bhende2008}, classification of power quality into 11 classes was done by class partitioning using the OAA technique. The MNNs were applied after feature extraction using the S-transform, and the different modules were integrated using a max activation function to produce the final output.

In \citet{Vlahogianni2007}, a subspatial domain partitioning, that is not class based, is used to train different modules on forecasting traffic volume at different road locations. In \citet{Aminian2007}, an electronic circuit is decomposed into multiple subcircuits to facilitate fault detection by several neural network modules. Feature domain partitioning is another form of manual partitioning, and is seen in \citet{Mendoza2009a,Mendoza2009b} where edge detection using a fuzzy inference system is done on target images to obtain edge vectors. Then different neural networks are trained on parts of these vectors, and the network outputs are combined using the Sugeno integral to produce the final classification results. Sometimes, the feature domain partitions are just different transformations of the data, with each transformation revealing different perspectives of the data. This is realised in \citet{Ciregan2012,Wang2015} where the input image together with its transformations via different image processing functions are used as inputs into a modular Convolutional Neural Network (CNN), in order to enhance classification performance.

\subsubsection{Learned}

Learned decomposition is the partitioning of the domain using a learning algorithm. Problem domains are often not easily separable. This is closely related to the problem of representation. If the domain can be manually decomposed into an optimal set of subdomains, that is a set of subspaces that capture all of the constraints of the problem compactly, this will significantly facilitate learning. However, usually the data generating process of the domain involves many interacting factors that are not readily observed. Problems like these typically require  learning algorithms to be applied both to the partitioning (explicitly or implicitly) and the overall classification problem.  

Learned decomposition facilitates the capturing of useful clustering patterns, especially complex ones that are not tractable by human designers. This intractability may stem from different sources like mathematical complexity or poorly understood problems. For example, the prediction of protein secondary and tertiary structure is often a complex and poorly understood process, that may take tremendous amounts of computational resources to simulate \citep{Freddolino2008,Allen2001}. However, learning algorithms often add computational cost to the overall process, since they typically involve adding an extra step for optimizing the model responsible for the decomposition.
  
Because of the clustering nature of learned decomposition, the mainstream approach involves applying unsupervised learning. In \citet{Ronen2002}, fuzzy clustering is used to dynamically partition the domain into regions, then a different multilayer perceptron (MLP) is trained on each region, and finally, the MLP outputs are integrated using a Sugeno integral like method. Also, in \citet{Fu2001}, a technique called Divide-and-Conquer Learning (DCL) is used to partition the domain whenever learning stalls. DCL acts by dividing training regions into easy and hard sets using Error Correlation Partitioning (ECP) \citep{Chiang1994}, which is based on optimizing a projection vector that separates data points according to their training error, then different modules are trained on each region and finally integrated using a gated network.

\subsection{Topology}

\begin{figure*}
\centering

\includegraphics[width=\textwidth]{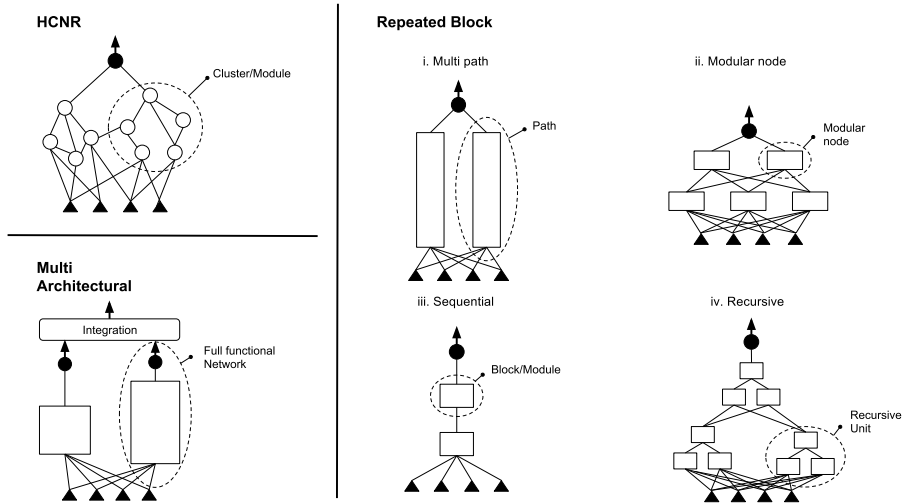}
\caption{Different modular neural network topologies.}
\label{fig:topol}

\end{figure*}

The topology of a network refers to how different nodes and modules within the network connect with each other, in order to produce the overall structure of the model. A neural network with a modular topology (\autoref{fig:topol}) exhibits a structure whereby nodes within a module are densely connected to each other, with sparse connectivity between modules. This is topological modularity, whereas functional modularity emerges when each topological module can be assigned a sub function of the whole task addressed by the neural network model. Topological modularity is a necessary, but not sufficient, condition of functional modularity. Without a learning algorithm that promotes functional modularity, topological modularity is not guaranteed to give rise to functional specialisation.
 
Neuroscience research sheds light on the modular topology of the nervous system. Neural circuitry in the brain is organized into modules at different levels of granularity, from cortical columns and neuronal nuclei to the whole anatomical areas of the brain's macro structure. It has been suggested in different works that the modularity of the brain arises from selection for evolvability and minimization of connection cost \citep{Bullmore2009,Clune2013}. 

Although the early inspiration for neural networks was the brain, artificial neural network research has mostly deviated from biological research. However, there are still occasional insights taken from biology (e.g. deep networks and convolutional structures, to name a few impactful examples) with the usual caveat that the aim is not to closely mimic the brain, but to solve real world problems in effective ways regardless of the source of the core ideas. So, although topological modularity is inspired by the brain's modular structure, it has metamorphosed into different forms that better suit the ANN domain.
 
The formation and learning of monolithic neural networks are hard problems. This is especially true with very deep neural networks. Deep learning faces several problems like overfitting, vanishing gradients and spatial crosstalk. Good topological modularization acts as a kind of regularization relative to highly connected monolithic networks. Some forms of modular topologies \citep{Larsson2016,He2016,Srivastava2015} provide shortcut paths for gradient flow which help to alleviate vanishing gradients. Moreover, the sparse connectivity of modules reduces spatial crosstalk.
 
One of the main problems with monolithic networks arises when something wrong occurs. In the vast majority of cases, neural networks are considered black box models. As such, it is usually unrealistically hard to decipher how a neural network makes its predictions. This stems mainly from a neural network's distributed representations, where nodes are tightly coupled, making separation of functions infeasible even in theory. This makes debugging and fixing deviations in behaviour very difficult. Topological modularity, especially if accompanied by functional modularity, can be exploited to localize functional errors so that more investigations may reveal possible solutions. In the case of full functional modularity, there are still distributed representations associated with different modules, however, since modules themselves are loosely coupled, this separation of concerns makes localizing deviation in some sense realistic.

\subsubsection{Highly-Clustered Non-Regular (HCNR)}

HCNR topology is a modular topology with non-regular and dense within-module connections, and sparse connectivity between different modules. Non-regularity here roughly means that the overall topology can’t be described by a template with repeating structures. This makes the topology generally hard to compress. Elements from graph theory can be used to formalize this notion using measures like characteristic path length and  clustering coefficient \citep{Watts1999}. Aside from high clustering, HCNR doesn’t have to exhibit properties such as the short average path length of small world (SW) networks. Thus, the broader category of HCNR includes small world topologies as special cases. 

Biology has shown significant interest in small world networks as increasing evidence suggests that many biological systems including genetic pathways, cellular signalling and brain wiring, exhibit small world topology. The evolutionary origins of the brain's modular structure is still controversial, but some hypotheses have  been suggested. In \citet{Kashtan2005}, it was suggested that evolution under a Modularly Varying Goals (MVG) environment yields modular structure. An MVG environment changes modularly, in the sense that the environmental goal varies through time, but each goal is comprised of the same common set of subgoals. A biological example is chemotaxis towards nutrients. The process of chemotaxis involves the same set of intermediate goals, like sensing, computing motion direction and moving, that are independent of the target nutrient. In another hypothesis \citep{Clune2013} it was suggested that modularity is bootstrapped by a pressure for minimizing costs pertaining to neuronal connections, and is then maintained by selection for evolvability. Random networks tend to have high connection cost due to dense connectivity, which is not the case in the brain's neural circuitry, which exhibits economical functional networks \citep{Achard2007}. Having a modular structure promotes evolvability, as accumulative evolutionary changes tend to be local in effect without disrupting other functions.

In the context of artificial neural networks, the sparse connectivity of HCNR and average short path of its special case SW, reduce computational complexity compared to monolithic networks, whilst maintaining information transfer efficiency. However, due to their structural complexity, analysing and adapting these types of networks to real world problems is hard and raises several technical difficulties. How many nodes should be in each module? Should module node counts vary? How much connectivity is allowed within a module? And what connection sparsity between modules is sparse enough? Also, formation of this network type is done either by modifying a regular lattice \citep{Bohland2001}, which is hard to adapt to all machine learning tasks, or via evolutionary algorithms \citep{Huizinga2014,Verbancsics2011,Garcia-Pedrajas2003,Mouret2009,Mouret2008}, which are lengthy and computationally expensive. The adoption of these two approaches, especially evolutionary algorithms, is a direct consequence of the previously mentioned difficulties and lack of good general engineering practices for HCNR.

The work done in \citet{Bohland2001} shows that an SW topology can approach the performance of random networks for associative memory tasks, with less connectivity than random networks, implying that associative memories can benefit from modularity. This network was constructed by rewiring a regular lattice randomly. The work by \citet{Bohland2001} shows that performance is not only about the quantitative nature of connectivity (i.e number of connections), but also about its qualitative nature (i.e how these connections are placed).

Evolutionary approaches for HCNR are either based on direct connection cost regularization, or the coevolution of modules. Both approaches tend to be biologically inspired, with connection cost regularization corresponding to the pressure of minimizing brain wiring, and coevolution inspired by species coevolution, where in this case each module is considered a different species. The evolutionary approach in \citet{Huizinga2014,Verbancsics2011} made use of the connection regularization studied in biological neural networks \citep{Clune2013,Bullmore2009} to promote HCNR modularity in the resulting model, which led to better performance on modular regular problems such as the retina problem. Another evolutionary approach relies on the cooperative coevolution model where modules are dependently co-evolved and their fitness is evaluated based on each module's performance and how well each module cooperates with other modules. COVNET \citep{Garcia-Pedrajas2003}, for example, achieved better results in some classification problems, like the Prima Indian and Cleveland Clinic Foundation Heart Disease datasets. COVNET also showed robustness to damage in some network parts.

\subsubsection{Repeated Block}

This topology of modular neural networks is essentially a structure of repeated units or building blocks connected in a certain configuration. The building blocks don’t have to be exact clones of each other, but they are assumed to share a general blueprint. The idea of global wiring schema in neural networks has its roots in biological studies and it is the underlying principle of the famous neuroevolution algorithm, HyperNEAT \citep{Stanley2009}. In \citet{Angelucci1997} it was shown that retinal projections in ferrets, normally relayed to the Lateral Geniculate Nucleus (LGN), when rewired to the Medial Geniculate Nucleus (MGN), normally a thalamic relay in the auditory pathway, led the MGN to develop eye-specific regions. Also, mammalian cortex is considered to be composed of repeating columnar structure \citep{Lodato2015}. These and other lines of evidence support the notion of a global mechanism of wiring and learning in the brain.
 
In the artificial realm, repeated block structure allows for easier analysis and extensibility of neural networks. On the theoretical level and due to the high regularity of these topologies, a very large structure can be described by a few equations. For example, a recursive structure like FractalNet \citep{Larsson2016}, can be described by a simple expansion rule. Also, due to regularity, scaling the capacity of these topologies tends to be very natural. We provide a taxonomy of repeated block topologies based on how the repeated units are wired together.

\paragraph{Multi-Path}

Multipath topology refers to neural networks with multiple semi-independent subnetworks connecting network inputs to outputs. In \citet{KienTuongPhan2015,Phan2016,Phan2017} this topology is named Parallel Circuits (PCs), which are inspired by the microcircuits in the retina \citep{Gollisch2010}. The retina is believed to be of significant computational importance to the visual pathway, not just a simple informational relay. In other words, it has been shown that the retina does perform complex computational tasks, such as motion analysis and contrast modulation, and delivers the results explicitly to downstream areas. Moreover, these microcircuits have been shown to exhibit some sort of multipath parallelism, embodied by semi-independent pathways involving different combinations of photoreceptor, horizontal, bipolar, amacrine and retinal ganglion cells. 

The separation of multiple paths allows for overall network parallelization, contrary to network expansion in terms of depth, where deeper layers depend on shallower ones, which makes parallelization problematic. Also, as in \citet{Ortin2005,Wang2015}, each path can be assigned to a different input modality which allows for modal integration. This resembles brain organization where different cortical areas process different modalities, and then different modalities get integrated by association areas. However, the introduction of multiple paths adds uncertainty in the form of new hyperparameters (e.g. numbers and widths of paths), which if to be determined empirically, often requires a phase of pre-optimization. Moreover, aside from obvious links to ensemble theory, as of yet there is no detailed theoretical justification for multiple paths. Why do empirical experiments show improved generalization performance \citep{Phan2016} of multipath over monolithic topologies? Does width (in terms of number of paths) promote problem decomposition just like depth promotes concept composition? To date there are no mature gradient-based learning algorithms to fully exploit the parallel circuit architecture, and which are likely to explicitly promote automatic task decomposition across paths.

In \citet{KienTuongPhan2015,Phan2016,Phan2017}, a multipath approach with shared inputs and outputs is shown to often exhibit better generalization than monolithic neural networks. Crucial to this improvement, was the development of a special dropout approach called DropCircuit, where whole circuits are probabilistically dropped during training. In another approach \citep{Guan2002}, called output parallelism, the inputs are shared between paths, while each path has a separate output layer. This technique can be applied when the output is easily decomposable. A very related approach can be found in \citet{Goltsev2015}, where a central common layer is connected to multiple paths, each used for a different output class. On the other hand, the work in \citet{Wang2015} enhances CNNs by allowing for two paths, one with the source image as input, and the other with a bilateral filtered version of it. In bilateral filtering, each pixel value is replaced by the average of its neighbours, taking into account the similarity between pixels, so that high frequency components can be suppressed while edges are preserved. The integration of images preprocessed in different ways facilitates the capturing of more useful features. This is motivated by the observation that convolution and pooling operations extract high frequency components, which causes simple shapes and less textured objects to gradually disappear. Bilateral filtering of one of the input images tends to suppress high frequency components, which allows the network to capture both simple and complex objects. The multi-path concept can be integrated with other topologies, as in \citet{Xie2016} where ResNetXt enhances ResNet's sequential topology by introducing modules that have multi-path structure.

\paragraph{Modular Node}

A modular node topology can be viewed as a normal monolithic feedforward neural network, where each node is replaced by a module consisting of multiple neurons. This expansion is computationally justified by replacing a single activation function depending on one weight vector, by a collection of functions or a function depending on multiple weight vectors. This has the effect of increasing the computational capability of the network, while maintaining a relatively small number of model parameters. Moreover, the regularity and sparsity of such a structure, combined sometimes with restricting weights to integer values, can be suitable for hardware realizations \citep{Sang-WooMoon2001}. On the other hand, this requires additional engineering decisions, like choosing the number of module neurons, how they are interconnected and what activation functions to use.

A special case of this topology is hierarchical modular topology, which consists of modules at different topological scales, where a higher level module is composed of submodules, each of which is composed of submodules, and so on. Hierarchical modularity is known to exist in brain networks \citep{Wang2011,Kaiser2010}, other biological networks and Very Large-Scale Integration (VLSI) electronic chips \citep{Meunier2010}. It has been argued that this form of modularity allows for embedding a complex topology in a low dimensional physical space.
 
In \citet{Sang-WooMoon2001,WeiJiang2007,PhyoPhyoSan2011} modular node topology is realized by replacing the nodes of a feedforward network by a two dimensional mesh of modules, with modular units each consisting of four neurons. The four neurons can be connected in four different configurations. This network, called Block-Based Neural Network (BBNN), was shown to be applicable to multiple tasks including pattern classification and robotic control, even when its weights were restricted to integer values. Another modular network called Modular Cellular Neural Network (MCNN) \citep{Karami2013} exhibits similar array like arrangement, where nine modules are arranged in a grid. A module in MCNN is composed of another grid of dynamic cells, where each cell is described by a differential equation. MCNNs were applied to texture segmentation successfully and benchmarked to other algorithms on the problem of edge detection. Another realization of this topology can be found in the Local Winner-Take-All (LWTA) network \citep{Srivastava2013}, where each node of a feedforward neural network is replaced by a block, each consisting of multiple non interconnected neurons. The network operates by allowing only the neuron with the highest activation in a block to fire, while suppressing other neurons. The block output is

\begin{equation} \label{eq:LWTA}
y_i = g(h_i^1, h_i^2, ..., h_i^n)
\end{equation}

where $g(.)$ is the local interaction function and $h_i^j$ is the activation of the $j$th neuron in block $i$. This is mainly inspired by the study of local competition in biological neural circuits. Network In a Network or (NIN) models \citep{Lin2013} are the modular node equivalents of CNNs, where each feature map is replaced by a micro MLP network, to allow for high-capacity non-linear feature mapping. The output of a single micro MLP is

\begin{equation} \label{eq:NIN}
\boldsymbol{f_{i,j}^l} = max(0, \boldsymbol{W_{l}f_{i,j}^{l-1}} + \boldsymbol{b_{l}})
\end{equation}

where $(i,j)$ are indices of the central pixel location, $l$ is the index of the MLP layer and $\boldsymbol{W}$ and $\boldsymbol{b}$ are the weights and bias, respectively. It is also interesting that the long short-term memory (LSTM) architecture \citep{Hochreiter1997}, the famous recurrent neural network (RNN), has a modular node structure, in which each node is an LSTM block. LSTM has shown state-of-the-art results in sequence modeling and different real-world problems \citep{Stollenga2015,Eyben2013,Soutner2013}. Moreover, Hierarchical Recurrent Neural Networks (HRNN), which are typically realised using LSTM or its simplification, the Gated Recurrent Unit (GRU), implements hierarchical modular topology, where the first hidden layer is applied to input sequentially and the layer output is generated every $n$ number of inputs, which is then propagated as input to the next layer and so on. Hence, the main difference between HRNNs and classic RNNs is that, for the former, hidden layer outputs are generated at evenly spaced time intervals larger than one. HRNNs have been used for captioning videos with a single sentence \citep{Pan2016} and with a multi-sentence paragraph \citep{Yu2016}, and for building end-to-end dialogue systems \citep{Serban2016}.

CapsNet was introduced in \citet{Sabour2017}, which is mainly a vision-centric neural network that attempts to overcome the limitations of CNNs. The main rationale behind CapsNet is representing objects using a vector of instantiation parameters that ensures equivariance with different object poses. CapsNet can be thought of as an ordinary CNN, in which each node is replaced by a vector-output module. The output of such a modular node in CapsNet is calculated as

\begin{equation} \label{eq:caps-squash}
\boldsymbol{v_j} = \frac{||\boldsymbol{s_j}||^2}{1 + \boldsymbol{||s_j||}^2}\frac{\boldsymbol{s_j}}{\boldsymbol{||s_j||}}
\end{equation}

where $\boldsymbol{s_j}$ is the node input and is calculated as
\begin{equation} \label{eq:caps-input}
\boldsymbol{s_j} = \sum_i{c_{ij}\boldsymbol{\hat{u}_{j|i}}},\  \boldsymbol{\hat{u}_{j|i}} = \boldsymbol{W_{ij}u_i} 
\end{equation}

where $\boldsymbol{u_i}$ is the output of a unit $i$ from the previous layer,  $\boldsymbol{W_{ij}}$ is a transformation matrix and $c_{ij}$ is a coupling coefficient. Coupling coefficients are calculated through a routing softmax as

\begin{equation} \label{eq:caps-softmax}
c_{ij} = \frac{e^{b_{ij}}}{\sum_k{e^{b_{ik}}}} 
\end{equation}

where $b_{ij}$ is a log prior probability which is initialized to zero and updated following the rule

\begin{equation} \label{eq:caps-update}
b_{ij} \leftarrow b_{ij} + \boldsymbol{\hat{u}_{j|i}}.\boldsymbol{v_j}
\end{equation}

This is called routing-by-agreement and acts to increase contributions from lower layer capsules that make good predictions regarding the state of a higher level capsule. 

\paragraph{Sequential}

Sequential topology consists of several similar units connected in series. The idea of composition of units has its roots in deep learning. Deep networks arise when multiple layers are connected in series. This allows for deep composition of concepts, where higher level representations are composed from lower level ones. The difference here is that the composed units consist of whole modules. But with added depth, convergence and generalization can become increasingly difficult, and one must therefore resort to tricks like dropout and batch normalization to make learning feasible. Moreover, there has been recent criticism of very deep networks based on the question of whether this extreme depth is really necessary \citep{Ba2014,Veit2016}, specially given that the brain can do more elaborate tasks with far fewer layers.

Inception networks \citep{Szegedy2015,Szegedy2016} and Xception networks \citep{Chollet2016} (built from an extreme version of an inception module), are essentially a sequential composition of multi-path convolutional modules. Highway networks were introduced in \citet{Srivastava2015} and can be seen as a sequentially connected block of modules, where each module output consists of the superposition of the layer output and layer input, weighted by two learning functions called the transform gate and carry gate respectively. The output of a single layer in a highway network can be modelled as 

\begin{equation} \label{eq:highway}
y = H(x, \boldsymbol{W_H}).T(x, \boldsymbol{W_T}) + x.C(x, \boldsymbol{W_C})
\end{equation}

where $H$ is the layer activation function, $T$ is the transform gate and $C$ is the carry gate. These two gates learn to adaptively mix the input and output of each module, acting as a regulator of information flow. A similar idea can be found in \citet{He2016} where a special case of highway networks called residual networks consists of the same structural unit but with both gates set to the identity function. This makes the residual layer output

\begin{equation} \label{eq:residual}
y = H(x, \boldsymbol{W_H}) + x
\end{equation}

This is motivated by simplifying the learning problem and enforcing residual function learning. Interestingly, the LSTM network, whose modular node topology was discussed above, exhibits a temporal form of sequential topology, where each LSTM block feeds its output to itself through time. So, expanding the LSTM block in time results in a temporal sequential topology, where the output of the LSTM block from the previous time step is considered as an input to the LSTM block in the current time step.

\paragraph{Recursive}

Networks with recursive topology exhibit nested levels of units, where each unit is defined by earlier units in the recursion. Usually, all of the units are defined by the same template of components wiring. Recursion has a long history and is considered to be a key concept in computer science. Although recursive problems can be solved without explicit recursion, the recursive solution is a more natural one. In theory an infinite structure can be defined in one analytical equation or using a simple expansion rule. Due to their recursive structure, networks with recursive topology are readily adaptable to recursive problems \citep{Franco2001}. Recursion also allows for very deep nesting while still permitting short paths, sometimes called information highways \citep{Larsson2016} that facilitate gradient back-propagation and learning. However, as mentioned earlier, excessive depth is criticised by some researchers and its necessity is becoming increasingly debatable.

FractalNet introduced in \citet{Larsson2016} exploits recursive topology to allow for very deeply nested structure that is relatively easy to learn despite its significant depth. It is inspired by the mathematically beautiful self-similar fractal, where going shallower or deeper in structure yields the same topology schema. It is defined as

\begin{equation} \label{eq:fractal}
f_{C+1}(x) = [(f_C \circ f_C)(x)] \oplus [conv(x)]
\end{equation}

where $C$ is the fractal index, $\circ$ means composition and $\oplus$ is a join operation. It is supposed that the effectively shorter paths for gradient propagation facilitate learning and protect against vanishing gradients. In \citet{Franco2001} the parity problem was decomposed recursively and a recursive modular structure was adapted for its solution. Also in this work on the parity problem, it was shown that generalization was systematically improved by degree of modularity, however, it was not obvious if that was a general conclusion applying to all problems.

\subsubsection{Multi-Architectural}

A multi-architectural topology consists of a combination of full network architectures, integrated together via a high-level and usually simple algorithm. Frequently, it is characterized by each component network having its separate output. The different architectures used may be similar (i.e homogeneous) or different (i.e heterogeneous). Architectural differences include, but are not limited to, differences in wiring scheme and activation functions. As different network architectures have different strengths and weaknesses (and make different errors), the integration is usually trying to exploit this diversity in order to achieve a more robust collective performance. Even when networks are similar, diversity can still be achieved since random initialization and stochastic learning makes each network converge differently. However, this usually entails training multiple architectures, which is time consuming and computationally expensive.
 
In \citet{Ciregan2012}, a homogeneous model is used where similar CNNs are trained on different types of pre-processing of the same image and their outputs are integrated by averaging. In \citet{Yu2018}, a referring expression is decomposed into three components, subject, location and relationship, where each component is processed using a separate visual attention module, which is essentially a CNN, and then the outputs of the different modules are combined. In \citet{Babaei2010}, a heterogeneous model consisting of two different RNNs, each modeling different protein structural information, is applied to predicting protein secondary structure. In \citet{Zhang2016}, a modular deep Q network is proposed to facilitate transferring of a learnt robotic control network from simulation to real environment. By modularising the network into three components, namely perception, bottleneck and control modules, the perception training can be done independently from the control task, while maintaining consistency through the bottleneck module acting as an interface between the two other modules. In \citet{Shetty2015,Yu2016,Pan2016} heterogeneous models of CNNs and RNNs are used for video captioning, where one or more CNNs are used for extracting features, which are used as inputs to RNNs for generating video captions. Generative Adversarial Networks (GANs) and their variants \citep{Kim2017} are also multi-architectural in nature. Another interesting example is the memory network \citep{Weston2014}, where multiple networks are composed end-to-end around a memory module to allow for the utilisation of past experiences. Essentially, a memory network is composed of a memory and four components, namely, $I$, $G$, $O$ and $R$. $I$ is the input network that translates the raw input into an internal representation $I(x)$. $G$ stands for generalization and it is responsible for updating memory based on the new input

\begin{equation} \label{eq:mem-g}
m_i = G(m_i, I(x), m)
\end{equation}

where $i$ is the index of the memory cell. After the memory is updated, another module, $O$, computes the output features based on the new input and the memory

\begin{equation} \label{eq:mem-o}
o = O(I(x), m)
\end{equation}

and finally, the $R$ module converts the output features into the desired format

\begin{equation} \label{eq:mem-r}
r = R(o)
\end{equation}

\subsection{Formation} \label{sec:form}

Formation refers to the technique used to construct the topology of the neural network. Manual formation involves expert design and trial and error. In manual formation, the human designer is guided by analytical knowledge, several heuristics and even crude intuition. Because of the difficulty and unreliability of manual formation, and a general lack of understanding of the relation between problems and the models they require, automatic techniques have been devised. Arguably the most popular automatic techniques are evolutionary algorithms, where the structure of the network is evolved over multiple generations, based on a fitness function that evaluates which individuals are more adapted. Another set of automatic formation algorithms constitute the learned formation category, where a learning algorithm is used not only for parameter (e.g. connection weight) optimization, but also for structure selection. Learned formation can be categorized into constructive and destructive algorithms \citep{Garcia-Pedrajas2003}. In constructive learned formation, the algorithm starts with a small model, learns until the performance stalls, adds components to expand capacity and iterates again. Destructive learned formation algorithms start with a big model that overfits, then iteratively remove nodes until the model generalizes well.
 
In order to form a modular neural network, one of these construction approaches needs to be modified in order to take modularization into account. With manual formation, it is in principle straightforward to modularize, where instead of designing a monolithic network, different modules are designed and combined to build an MNN. On the other hand, while standard evolutionary algorithms can produce modular structure, they are usually modified using techniques like cooperative coevolution, given that the latter are generally seen to be more effective for evolving modular structure. In the case of learned formation, learning algorithms usually take modularity explicitly into account. So, the machine learning task becomes that of learning both modular structure and the parameters (e.g. weights) of that structure. A variant of learned formation, which we call implicit learned formation, is a learning algorithm that is implicitly sampling or averaging from a set of modules, so that the overall effective structure of the network can be seen as a modular one.

\subsubsection{Manual}

In manual formation, modular networks are built by manual design and composition of different modules. This type of formation provides useful opportunities for integrating good engineering principles and prior knowledge of the target problem into the modular neural network. For example, in \citet{Babaei2010}, the system for predicting protein secondary structure is formed from two RNN modules that model two different aspects of the process, namely, short and long range interactions. Fine control over what to include or exclude from the model can lead to a robust combination of well performing components. However, regardless of how this sounds theoretically plausible, limited understanding of the underlying structures of most real-world problems and limited, to date, research into good neural modularization practices, make this hard in practice.

In \citet{DeNardi}, different modules are manually composed together to implement a helicopter control system, based on the practices of human designed Proportional–Integral–Derivative (PID) controllers. The PID components are replaced progressively by their neural network counterparts, until the whole control is done by the MNN. More formally in \citet{Guang-BinHuang2003}, analytical introduction of modular layers into feedforward neural networks allows for reducing the number of nodes required to learn a target task. This is a case that shows how good engineering could be integrated, through formal analysis, into the formation of modular neural networks. 

\subsubsection{Evolutionary}

Evolutionary algorithms represent the current state of the art in formation methods for modular neural networks. This is clearly biologically inspired by the neuroevolution of the brain, which is shown to be highly modular both in topology and functionality \citep{Aguirre2002,Bullmore2009}. Aside from biological inspiration, evolving modular structure has gained momentum as an effective approach to modularity formation, partly because of a lack of fundamental learning principles supporting artificial neural modularity. Adapting evolution to the problem of modularity formation, through connection cost regularization \citep{Huizinga2014} or cooperative coevolution \citep{Garcia-Pedrajas2003}, partly delegates the problem of choosing modularity-related hyperparameters, such as the number and structure of modules, or connection schema, to a fitness function. Furthermore, evolutionary algorithms are the only fitness-based approach to producing HCNR topology, whereas other methods rely on random modifications to regular networks. On the down side, as already mentioned, evolutionary algorithms tend to be computationally expensive.
 
In \citet{Garcia-Pedrajas2003}, COVNET was introduced, which is a modular network formed using a cooperative coevolutionary algorithm. Every module is called a nodule, and is defined as a set of neurons that are allowed to connect to each other and to input/output nodes, but are not allowed to connect to neurons in other nodules. Every nodule is selected from a genetically separated population and different nodules are combined together to form individuals of the network population. To achieve cooperative coevolution, it is not sufficient to assign fitness values to networks, but it is also necessary to assign fitness values to nodules. The fitness of a network is straightforward, where obviously it corresponds to how well the network performs on its target task. The fitness assignment of nodules must enforce: (1) competition, so that different subpopulations don’t converge to the exact same function, (2) cooperation, so different subpopulations develop complementary features and (3) meaningful contribution of a nodule to network performance, such that poorly contributing nodules are penalized. In COVNET, a combination of different measures is used to satisfy these criteria. Substitution is used to promote competition, where the best $k$ networks are selected and a nodule $a$ is replaced by a nodule $b$ from the same subpopulation; then the networks fitnesses are recalculated, and nodule $a$ is assigned fitness proportional to the average difference between the network fitnesses with nodule $a$ and the network fitnesses with the substitution nodule $b$. Difference is used to promote cooperation between nodules by promoting competition between nodule subpopulations, so that they don't develop the same behaviour. Difference is done by eliminating a nodule $a$ from all the networks where it is present, then recalculating network fitnesses; then the nodule is assigned fitness proportional to the average difference between fitnesses of the networks with the nodule and the networks without it. Finally, best $k$ is used to assess the meaningful contribution of a nodule, where nodule fitness is proportional to the mean of the fitnesses of the best $k$ networks. This has the effect of rewarding nodules in the best performing networks, whilst not penalizing a good nodule in a poor performing network.

Promoting modularity through connection cost minimization \citep{Huizinga2014} is biologically inspired as evidence suggests that the evolution of the brain, guided by the minimization of wiring length, besides improving information transfer efficiency, produces modular structure \citep{Clune2013,Bullmore2009}. In \citet{Husken2002}, modularity emerges through evolution by selection pressure for both fast and accurate learning. In \citet{Di2001}, a modular multi-path neural network was evolved for solving the what and where task of identifying and localizing objects using a neural network. In another type of approach, modules are used as substrates for evolutionary algorithms. For example, in \citet{Braylan2015} pre-learned networks (modules) were combined using evolutionary algorithms, in an attempt to implement knowledge transfer. In \citet{Calabretta2000}, evolution was implemented using a technique called duplication-based modular architecture, where the architecture can grow in the number of modules by mutating a set of special duplicating genes. In \citet{Miikkulainen2017} a population of blueprints, each represented by a graph of module pointers, was evolved using CoDeepNEAT, alongside another population of modules, evolved using DeepNEAT, an algorithm based on NEAT \citep{Stanley2002}, to develop deep modular neural networks. CoDeepNEAT seems to be a generalization of a previous algorithm called ModularNEAT \citep{Reisinger2004}, where modules are evolved using classic NEAT and blueprints are shallow specifications of how to bind modules to the final network input and output.

In \citet{Fernando2017}, an interesting approach to evolutionary formation is introduced, where only some pathways in a large neural network composed of different modules are trained at a given time. The aim is to achieve multi-task learning. The pathways are selected through a genetic algorithm that uses a binary tournament to choose two pathways through the network. These pathways are then trained for a number of epochs to evaluate their fitness. The winner genome overwrites the other one and gets mutated before repeating the tournament. At the end of the training for some task, the fittest pathway is fixed and the process is repeated for the next task.

\subsubsection{Learned}

Learned formation is the usage of learning algorithms to induce modular structure in neural networks. Learned formation attempts to integrate structural learning into the learning phase, such that the learning algorithm affects network topology as well as parameters. We identified two variants of learned formation in the literature. Explicit learned formation uses machine learning algorithms to promote modularity, predict the structure of modular neural networks and specify how modules should be wired together. On the other hand, implicit learned formation corresponds to learning algorithms that implicitly sample from multiple modules during training, although during the prediction phase, the network is explicitly monolithic whilst effectively simulating a modular network. Learned formation, just like evolutionary formation, allows for dynamic formation of modules. Moreover, as mentioned above, it can effectively sample from a large set of models, which is why it is often referred to as effectively implementing ensemble averaging \citep{Srivastava2014, Huang2016, Singh2016, Larsson2016}. The main disadvantage for dynamic algorithms like these is added computational overhead. Also, for implicit learned formation, the network is still densely connected and therefore computationally expensive, and modules are generally sampled randomly without any preference for better modules.

In \citet{Andreas2016a,Andreas2016b,Hu2016}, which exemplifies recent work on explicit learned formation, the problem of relating natural language to images was addressed. A set of modular blocks, each specialised in a certain function (e.g. attention and classification), were used as building units of a modular neural network, where different dynamic techniques were applied to assemble units together into an MNN that was capable of answering complex questions about images and comprehending referential expressions. For example, in \citet{Andreas2016b}, two distributions were defined: (1) layout distributions, defined over possible layouts/structures of the network and, (2) execution model distributions, defined over selected model outputs. The overall training was done end-to-end with reinforcement learning.

One of the most well known implicit learned formation techniques is dropout \citep{Srivastava2014}. Dropout acts by dropping random subsets of nodes during learning, as a form of regularization that prevents interdependency between nodes. Dropout is effectively sampling from a large space of available topologies during learning, because each learning iteration acts on a randomly carved sparse topology. In the prediction phase, networks are effectively averaging those random topologies to produce the output. Stochastic depth \citep{Huang2016} is another dropping technique used in training  residual networks, which acts by dropping the contribution of whole layers. Swapout \citep{Singh2016} generalizes dropout and stochastic depth, such that it is effectively sampling from a larger topological space, where a combination of dropping single units and whole layers is possible. DropCircuit \citep{Phan2016,Phan2017} is another related technique, which is an adaptation of dropout to a particular type of multipath neural network called Parallel Circuits. In this technique, whole paths are randomly dropped, such that learning iterations are acting on random modular topologies. This effectively corresponds to sampling from a topological space of modular neural networks. \citet{Blundell2015} introduced Bayes by Backprop, a learning algorithm that approximates Bayesian inference which is, as applied to a neural network, the sum of the predictions made by different weight configurations, weighted by their posterior probabilities. This is essentially an ensemble of an infinite number of neural networks. As this form of expectation is intractable, it is approximated using variational learning, using different tricks such as Monte Carlo approximation and a scale mixture prior.

\subsection{Integration} \label{sec:integ}

Integration is how different module outputs are combined to produce the final output of the MNN. Integration may be cooperative or competitive. In cooperative integration, all the MNN modules, contribute to the integrated output. On the other hand , competitive integration selects only one module to produce the final output. The perspective of integration is different from that of formation, where the latter is concerned with the processes that gives rise to modular structure, and the former is concerned with the structures and/or algorithms that use different modules in order to produce model outputs. Integration is a biologically inspired theme of brain structure, where hierarchical modular structures work together to solve a continually changing set of complex and interacting environmental goals.

\subsubsection{Arithmetic-Logic}

Arithmetic-Logic (AL) integration corresponds to a set of techniques that combine different modules through a well-defined algorithmic procedure, combining mathematical operators and logic. For problems that can be described using a sequence of algorithmic steps, this is the simplest and most straightforward approach, and is the most natural hook for integrating prior knowledge. It is worth mentioning that while the relation between steps needs to be algorithmically defined, the computation of the steps themselves is not necessarily well-defined. For example, a car control system may want to steer away from an obstacle once it has identified one. The relation between identification and steering away is AL-defined, while the identification of obstacles is not generally algorithmically defined. Moreover, AL integration allows for module decoupling, where each module has its well-defined interpretable output, which further makes debugging easy. However, in machine learning tasks, due to our limited understanding of problem domains and corresponding data generating processes, it is rarely the case that problems can easily be decomposed into AL steps.

In \citet{Gradojevic2009}, multiple neural networks were logically integrated, where each network was trained on only a part of the input space, and the output was integrated at the prediction phase by selecting the network that corresponded to the input subspace. This is a competitive integration type scheme. A more complex integration was done in \citet{DeNardi}, where neural network components for a helicopter control system were cooperatively integrated based on the AL of a hand designed PID controller. In \citet{Wang2012}, two CNNs, one being a text detector and the other a character recognizer, were logically integrated, where the detector determined image locations containing text and the recognizer extracted text given these locations. In \citet{Eppel2017}, the recognition of the parts of an object was done in two steps, where in the first step, a CNN was used to segment the image to separate the object from its background, then in the second step another CNN was applied to the original image and the segmentation map to identify the object parts.

\subsubsection{Learned}

Learned integration consists of the composition of modules through a learning algorithm. Here, learning is concerned with how to optimally combine modules in order to obtain the best possible performance on the target problem. Composing modules to solve a certain problem is not straightforward, involving complex interactions between modules. Using learning algorithms in modular integration helps to capture useful complex relationships between modules. Even when subproblems are readily composable into a final solution, learning algorithms can find shortcuts that can help formulate more efficient solutions. However, the introduction of learning can result in unnecessary computational overhead, and can give rise to tightly coupled modules, often leading to overfitting and harder debugging. A very common type of learned integration is synaptic integration, where different modules are combined together by converging to a common parametric layer, which determines, through learning, the contribution of each module to the final output.
 
In \citet{Almasri2005}, several neural network outputs were integrated together, to predict nitrate distribution in ground water, using a gating network. A gating network is a very common integration technique, where a specialised network is used to predict a vector of weights, which is used to combine the outputs of different experts (i.e networks). In \citet{Zheng2006}, a Bayesian probability model was used to combine MLP and RBF network predictions based on how well each module performed in previous examples. This Bayesian model tended to give more weight to the module that performed better on previous examples in a certain target period of prediction. Fuzzy logic has also been used as a tool for learned integration \citep{Melin2007,Hidalgo2009}. In \citet{Mendoza2009a,Mendoza2009b,Melin2011} an image recognition MNN was proposed where different neural networks were trained on part of the edge vector extracted from the target image. Fuzzy logic was then used to integrate different neural network outputs by assessing the relevance of each module using a fuzzy inference system and integrating using the Sugeno integral. Synaptic integration was done in \citet{Anderson2016}, with the aim of achieving transfer learning on a small amount of data. A set of untrained modules were added to a pretrained network and integrated by learning while freezing the original network weights. In another work \citep{Terekhov2015}, a similar approach utilized synaptic integration for multi-task learning, where an initial network was trained on some task, followed by a modular block of neurons that were added and integrated by training on a different task, while again freezing the original network parameters.

\onecolumn

\fontsize{6}{6}\selectfont

\renewcommand{\arraystretch}{1.08}

\tablefirsthead{%
   \hline
    \textbf{Technique} & \textbf{Advantages} & \textbf{Disadvantages}\\
   \hline}
 \tablehead{%
   \hline
   \multicolumn{3}{l}{\small\sl continued from previous page}\\
   \hline
    \textbf{Technique} & \textbf{Advantages} & \textbf{Disadvantages}\\
   \hline}
 \tabletail{%
   \hline
   \multicolumn{3}{l}{\small\sl continued on next page}\\
   \hline}
 \tablelasttail{\hline}
 \topcaption{Advantages and disadvantages of different technique}
\label{tab:adv-disadv}

\begin{center}

\begin{xtabular}{l p{.4\linewidth} p{.4\linewidth} }

\hline

\multicolumn{3}{l}{\textbf{A. Domain}}\\ \hline

1. Manual &

\begin{itemize}[noitemsep,topsep=0pt]
\item Prior knowledge integration
\item Fine control over partitions
\end{itemize} &

\begin{itemize}[noitemsep,topsep=0pt]
\item Partitions are hard to define
\item Relation between decomposition and solution is not straightforward
\item Separation of variation factors is hard
\end{itemize}\\

\multicolumn{3}{p{\textwidth}}{e.g \citet{Anand1995,Oh2002,Rudasi1991,Subirats2010,Bhende2008,Mendoza2009a,Mendoza2009b,Ciregan2012,Wang2015,Vlahogianni2007,Aminian2007}}\\ 
\hline
% ----------------------------
2. Learned &

\begin{itemize}[noitemsep,topsep=0pt]
\item Capture useful relations not tractable by human designer
\end{itemize} &

\begin{itemize}[noitemsep,topsep=0pt]
\item Computational cost and extra step of learning the decomposing model
\end{itemize}\\

\multicolumn{3}{p{\textwidth}}{e.g \citet{Ronen2002,Fu2001,Chiang1994}}\\ 
\hline
% ----------------------------
\multicolumn{3}{l}{\textbf{B. Topology}}\\ \hline
% ---------------------------
1. HCNR &
\begin{itemize}[noitemsep,topsep=0pt]
\item Sparse connectivity
\item Short average path
\end{itemize} &

\begin{itemize}[noitemsep,topsep=0pt]
\item Complex structure
\item Hard to analyse and adapt to problems
\item Formation difficulty
\end{itemize}\\

\multicolumn{3}{p{\textwidth}}{e.g \citet{Bohland2001,Huizinga2014,Verbancsics2011,Garcia-Pedrajas2003,Mouret2009,Mouret2008}}\\ 
\hline
% --------------------------
\multicolumn{3}{l}{2. Repeated Block}\\ \hline
% --------------------------
2.1. Multi-Path &

\begin{itemize}[noitemsep,topsep=0pt]
\item Parallelizable
\item Suitable for multi-modal  integration
\end{itemize} &

\begin{itemize}[noitemsep,topsep=0pt]
\item Additional hyperparameters
\item Currently lacks theoretical justification
\end{itemize}\\

\multicolumn{3}{p{\textwidth}}{e.g \citet{KienTuongPhan2015,Phan2017,Ortin2005,Wang2015,Phan2016,Xie2016,Guan2002}}\\ 
\hline
% -------------------------
2.2. Modular Node &

\begin{itemize}[noitemsep,topsep=0pt]
\item Computational capability with relatively fewer parameters
\item Can be adapted for hardware implementation
\end{itemize} &

\begin{itemize}[noitemsep,topsep=0pt]
\item Additional hyperparameters
\end{itemize}\\

\multicolumn{3}{p{\textwidth}}{e.g \citet{Sang-WooMoon2001,WeiJiang2007,Serban2016,Soutner2013,PhyoPhyoSan2011,Karami2013,Pan2016,Srivastava2013,Lin2013,Eyben2013,Yu2016,Hochreiter1997,Stollenga2015,Wang2011,Kaiser2010}}\\ 
\hline
% ------------------------
2.3. Sequential &

\begin{itemize}[noitemsep,topsep=0pt]
\item Deep composition
\end{itemize} &

\begin{itemize}[noitemsep,topsep=0pt]
\item Hard training
\item Excessive depth is arguably unnecessary
\end{itemize}\\

\multicolumn{3}{p{\textwidth}}{e.g \citet{Szegedy2015,Szegedy2016,Chollet2016,Srivastava2015,He2016}}\\ 
\hline
% ------------------------
2.4. Recursive &

\begin{itemize}[noitemsep,topsep=0pt]
\item Readily Adaptable to recursive problems
\item Deep nesting with short paths
\end{itemize} &

\begin{itemize}[noitemsep,topsep=0pt]
\item Excessive depth is arguably unnecessary
\end{itemize}\\

\multicolumn{3}{p{\textwidth}}{e.g \citet{Franco2001,Larsson2016}}\\ 
\hline

3. Multi-Architectural &

\begin{itemize}[noitemsep,topsep=0pt]
\item Better collective performance
\item Error tolerance
\end{itemize} &

\begin{itemize}[noitemsep,topsep=0pt]
\item Computationally complex
\end{itemize}\\

\multicolumn{3}{p{\textwidth}}{e.g \citet{Ciregan2012,Babaei2010,Shetty2015,Yu2016,Pan2016,Kim2017,Weston2014}}\\ 
\hline
% -----------------------------
\multicolumn{3}{l}{\textbf{C. Formation}}\\ \hline
% ----------------------
1. Manual &

\begin{itemize}[noitemsep,topsep=0pt]
\item Prior knowledge integration
\item Fine control over components
\end{itemize} &

\begin{itemize}[noitemsep,topsep=0pt]
\item Hard in practice
\end{itemize}\\

\multicolumn{3}{p{\textwidth}}{e.g \citet{DeNardi,Guang-BinHuang2003}}\\ 
\hline
% ----------------------
2. Evolutionary &

\begin{itemize}[noitemsep,topsep=0pt]
\item Adaptable way for modularity formation
\item Suitable for HCNR formation
\end{itemize} &

\begin{itemize}[noitemsep,topsep=0pt]
\item Lengthy and computationally complex
\end{itemize}\\

\multicolumn{3}{p{\textwidth}}{e.g \citet{Huizinga2014,Garcia-Pedrajas2003,Braylan2015,Miikkulainen2017,Reisinger2004,Husken2002,Calabretta2000,Di2001}}\\ 
\hline
% ---------------------
3. Learned &

\begin{itemize}[noitemsep,topsep=0pt]
\item Dynamic formation of modularity
\item Sample from large set of models
\end{itemize} &

\begin{itemize}[noitemsep,topsep=0pt]
\item Computational complexity
\item In implicit learned variant, networks are densely connected
\end{itemize}\\

\multicolumn{3}{p{\textwidth}}{e.g \citet{Srivastava2014,Huang2016,Singh2016,Larsson2016,Andreas2016a,Andreas2016b,Hu2016,Phan2016,Phan2017,Blundell2015}}\\ 
\hline
% ---------------------
\multicolumn{3}{l}{\textbf{D. Integration}}\\ \hline
% ---------------------
1. Arithmetic-Logic &

\begin{itemize}[noitemsep,topsep=0pt]
\item Prior knowledge integration
\item Loosely coupled modules
\end{itemize} &

\begin{itemize}[noitemsep,topsep=0pt]
\item Difficult in practice
\end{itemize}\\

\multicolumn{3}{p{\textwidth}}{e.g \citet{Gradojevic2009,DeNardi,Wang2012}}\\ 
\hline
% --------------------
2. Learned &

\begin{itemize}[noitemsep,topsep=0pt]
\item Captures complex relations
\end{itemize} &

\begin{itemize}[noitemsep,topsep=0pt]
\item Computationally complex
\item Tightly coupled modules
\end{itemize}\\

\multicolumn{3}{p{\textwidth}}{e.g \citet{Zheng2006,Mendoza2009b,Mendoza2009a,Almasri2005,Melin2007,Melin2011,Hidalgo2009}}\\ 
\hline
% --------------------

\end{xtabular}
\end{center}

\normalsize

\ifthenelse{\boolean{togglecolumn}}{\twocolumn}{}

\section{Conclusion}

This review aimed at introducing and analysing the main modularization techniques used in the field of neural networks so far, in an attempt to provide researchers and practitioners with insights on how to systematically implement neural modularity in order to harness its advantages. We devised a taxonomy of modularization techniques with four main categories, based on the target of the modularization process, i.e.: domain, topology, formation and integration. We further divided each category into subcategories based on the nature of the techniques involved. We discussed the advantages and disadvantages of the techniques, and how they are used to solve real-world problems. Analysis and empirical results show that modularity can be advantageous over monolithic networks in many situations.
 
The review has shown that a wide variety of algorithms for modularization exists, acting on different parts of the MNN life cycle. We have shown that advances in MNNs are not restricted to biologically inspired topological modularity. The quest for modularity in ANNs is far from being a mere case of enforcing networks to be partial replicas of the brain. Even topological modularity is often a vague imitation of brain structure. As the ANN literature has increasingly diverged from its early biological roots, so has modularity metamorphosed into different shapes and techniques, ranging from biologically-inspired to purely engineering practices.

The techniques reviewed here have ranged from explicit expert-knowledge based techniques to fully automated implicit modularization techniques, each having its specific set of pros and cons and suitability for particular problems. Some techniques were found to be tailored to satisfy the specific constraints of particular problems, while others were found to be generic, trading specialization performance for full automation and generalizability. Neural modularization was shown to be a sequential application of techniques, which we called modularization chain, where each technique acts on a different aspect of the neural network. 
 
Although, as discussed, modularity has many advantages over monolithic deep networks, the main trend is still oriented towards monolithic deep neural networks. This is mainly due to the many successes of monolithic deep learning in different areas throughout the last decade. Also, the lack of extensive research into learning and formation techniques for neural modularity makes it hard for practitioners to efficiently deploy the approach. Contrary to this, monolithic networks have attracted extensive research that has generated a critical mass of theoretical insights and practical tricks, which facilitate their deployment. Evolutionary algorithms are currently the main actors in complex modular neural network construction. However, the debate of whether evolutionary algorithms are the best approach for MNN formation and if they harness the full power of modularization and problem decomposition is still open. Also, there is still a significant gap on how to stimulate problem decomposition in modular networks, so that their topological modularity may also become a full functional modularity.

We tentatively predict that as the challenges facing deep learning become increasingly hard, a saturation phase will eventually be reached where depth and learning tricks may not be enough to fuel progress in deep learning. We don’t view modularity as a replacement for depth, but as a complementary and integrable approach to deep learning, especially given that excessive depth is becoming increasingly criticized for reasons of computational cost and extraneousness. The dilemma is similar to the software quality problem, where exponential growth in hardware efficiency is masking poor algorithmic optimization. We believe that as deep learning becomes increasingly applied to more challenging and general problems, the need for robust Artificial General Intelligence practices will sooner or later promote the modularization of neural networks.

\begin{acknowledgements}
This is a pre-print of an article published in Artificial Intelligence Review. The final authenticated version is available online at:  https://doi.org/10.1007/s10462-019-09706-7
\end{acknowledgements}

\bibliographystyle{spbasic}      % basic style, author-year citations
\bibliography{main}   % name your BibTeX data base

\begin{thebibliography}{125}
\providecommand{\natexlab}[1]{#1}
\providecommand{\url}[1]{{#1}}
\providecommand{\urlprefix}{URL }
\expandafter\ifx\csname urlstyle\endcsname\relax
  \providecommand{\doi}[1]{DOI~\discretionary{}{}{}#1}\else
  \providecommand{\doi}{DOI~\discretionary{}{}{}\begingroup
  \urlstyle{rm}\Url}\fi
\providecommand{\eprint}[2][]{\url{#2}}

\bibitem[{Achard and Bullmore(2007)}]{Achard2007}
Achard S, Bullmore E (2007) {Efficiency and cost of economical brain functional
  networks}. PLoS Computational Biology 3(2):0174--0183,
  \doi{10.1371/journal.pcbi.0030017}

\bibitem[{Aguirre et~al.(2002)Aguirre, Huerta, Corbacho, and
  Pascual}]{Aguirre2002}
Aguirre C, Huerta R, Corbacho F, Pascual P (2002) Analysis of biologically
  inspired small-world networks. In: International Conference on Artificial
  Neural Networks, Springer, pp 27--32

\bibitem[{Allen et~al.(2001)Allen, Almasi, Andreoni, Beece, Berne, Bright,
  Brunheroto, Cascaval, Castanos, Coteus, Crumley, Curioni, Denneau, Donath,
  Eleftheriou, Flitch, Fleischer, Georgiou, Germain, Giampapa, Gresh, Gupta,
  Haring, Ho, Hochschild, Hummel, Jonas, Lieber, Martyna, Maturu, Moreira,
  Newns, Newton, Philhower, Picunko, Pitera, Pitman, Rand, Royyuru, Salapura,
  Sanomiya, Shah, Sham, Singh, Snir, Suits, Swetz, Swope, Vishnumurthy, Ward,
  Warren, and Zhou}]{Allen2001}
Allen F, Almasi G, Andreoni W, Beece D, Berne BJ, Bright A, Brunheroto J,
  Cascaval C, Castanos J, Coteus P, Crumley P, Curioni A, Denneau M, Donath W,
  Eleftheriou M, Flitch B, Fleischer B, Georgiou CJ, Germain R, Giampapa M,
  Gresh D, Gupta M, Haring R, Ho H, Hochschild P, Hummel S, Jonas T, Lieber D,
  Martyna G, Maturu K, Moreira J, Newns D, Newton M, Philhower R, Picunko T,
  Pitera J, Pitman M, Rand R, Royyuru A, Salapura V, Sanomiya A, Shah R, Sham
  Y, Singh S, Snir M, Suits F, Swetz R, Swope WC, Vishnumurthy N, Ward TJC,
  Warren H, Zhou R (2001) {Blue Gene: A vision for protein science using a
  petaflop supercomputer}. IBM Systems Journal 40(2):310--327,
  \doi{10.1147/sj.402.0310},
  \urlprefix\url{http://ieeexplore.ieee.org/document/5386970/}

\bibitem[{Almasri and Kaluarachchi(2005)}]{Almasri2005}
Almasri MN, Kaluarachchi JJ (2005) Modular neural networks to predict the
  nitrate distribution in ground water using the on-ground nitrogen loading and
  recharge data. Environmental Modelling \& Software 20(7):851--871

\bibitem[{Aminian and Aminian(2007)}]{Aminian2007}
Aminian M, Aminian F (2007) A modular fault-diagnostic system for analog
  electronic circuits using neural networks with wavelet transform as a
  preprocessor. IEEE Transactions on Instrumentation and Measurement
  56(5):1546--1554

\bibitem[{Anand et~al.(1995)Anand, Mehrotra, Mohan, and Ranka}]{Anand1995}
Anand R, Mehrotra K, Mohan C, Ranka S (1995) {Efficient classification for
  multiclass problems using modular neural networks}. IEEE Transactions on
  Neural Networks 6(1):117--124, \doi{10.1109/72.363444}

\bibitem[{Anderson et~al.(2016)Anderson, Shaffer, Yankov, Corley, and
  Hodas}]{Anderson2016}
Anderson A, Shaffer K, Yankov A, Corley CD, Hodas NO (2016) {Beyond Fine
  Tuning: A Modular Approach to Learning on Small Data}
  \urlprefix\url{https://arxiv.org/abs/1611.01714v1}, \eprint{1611.01714}

\bibitem[{Andreas et~al.(2016a)Andreas, Rohrbach, Darrell, and
  Klein}]{Andreas2016a}
Andreas J, Rohrbach M, Darrell T, Klein D (2016a) Neural module networks. In:
  Proceedings of the IEEE Conference on Computer Vision and Pattern
  Recognition, pp 39--48

\bibitem[{Andreas et~al.(2016b)Andreas, Rohrbach, Darrell, and
  Klein}]{Andreas2016b}
Andreas J, Rohrbach M, Darrell T, Klein D (2016b) {Learning to Compose Neural
  Networks for Question Answering} \eprint{1601.01705}

\bibitem[{Angelucci et~al.(1997)Angelucci, Clasc{\'{a}}, Bricolo, Cramer, and
  Sur}]{Angelucci1997}
Angelucci a, Clasc{\'{a}} F, Bricolo E, Cramer KS, Sur M (1997) {Experimentally
  induced retinal projections to the ferret auditory thalamus: development of
  clustered eye-specific patterns in a novel target.} The Journal of
  neuroscience : the official journal of the Society for Neuroscience
  17(6):2040--2055

\bibitem[{Auda and Kamel(1998)}]{Auda1998}
Auda G, Kamel M (1998) {Modular Neural Network Classifiers: A Comparative
  Study}. Journal of Intelligent and Robotic Systems 21:117--129,
  \doi{10.1023/A:1007925203918}

\bibitem[{Auda and Kamel(1999)}]{Auda1999}
Auda G, Kamel M (1999) {Modular neural networks: a survey.} International
  journal of neural systems 9(2):129--51

\bibitem[{Ba and Caruana(2014)}]{Ba2014}
Ba J, Caruana R (2014) Do deep nets really need to be deep? In: Advances in
  neural information processing systems, pp 2654--2662

\bibitem[{Babaei et~al.(2010)Babaei, Geranmayeh, and Seyyedsalehi}]{Babaei2010}
Babaei S, Geranmayeh A, Seyyedsalehi SA (2010) {Protein secondary structure
  prediction using modular reciprocal bidirectional recurrent neural networks}.
  Computer Methods and Programs in Biomedicine 100(3):237--247,
  \doi{10.1016/j.cmpb.2010.04.005}

\bibitem[{Bengio et~al.(2015)Bengio, Vinyals, Jaitly, and Shazeer}]{Bengio2015}
Bengio S, Vinyals O, Jaitly N, Shazeer N (2015) {Scheduled Sampling for
  Sequence Prediction with Recurrent Neural Networks}.
  \urlprefix\url{http://papers.nips.cc/paper/5956-scheduled-sampling-for-sequence-prediction-with-recurrent-neural-networks}

\bibitem[{Bengio et~al.(2009)Bengio, Louradour, Collobert, and
  Weston}]{Bengio2009}
Bengio Y, Louradour J, Collobert R, Weston J (2009) {Curriculum learning}. In:
  Proceedings of the 26th Annual International Conference on Machine Learning -
  ICML '09, ACM Press, New York, New York, USA, pp 1--8,
  \doi{10.1145/1553374.1553380},
  \urlprefix\url{http://portal.acm.org/citation.cfm?doid=1553374.1553380}

\bibitem[{Bhende et~al.(2008)Bhende, Mishra, and Panigrahi}]{Bhende2008}
Bhende C, Mishra S, Panigrahi B (2008) {Detection and classification of power
  quality disturbances using S-transform and modular neural network}. Electric
  Power Systems Research 78(1):122--128, \doi{10.1016/j.epsr.2006.12.011}

\bibitem[{Blundell et~al.(2015)Blundell, Cornebise, Kavukcuoglu, and
  Wierstra}]{Blundell2015}
Blundell C, Cornebise J, Kavukcuoglu K, Wierstra D (2015) Weight uncertainty in
  neural networks. arXiv preprint arXiv:150505424

\bibitem[{Bohland and Minai(2001)}]{Bohland2001}
Bohland JW, Minai AA (2001) {Efficient associative memory using small-world
  architecture}. Neurocomputing 38:489--496,
  \doi{10.1016/S0925-2312(01)00378-2}

\bibitem[{Brandes et~al.(2008)Brandes, Delling, Gaertler, Gorke, Hoefer,
  Nikoloski, and Wagner}]{Brandes2008}
Brandes U, Delling D, Gaertler M, Gorke R, Hoefer M, Nikoloski Z, Wagner D
  (2008) {On modularity clustering}. IEEE Transactions on Knowledge and Data
  Engineering 20(2):172--188, \doi{10.1109/TKDE.2007.190689}

\bibitem[{Braylan et~al.(2015)Braylan, Hollenbeck, Meyerson, and
  Miikkulainen}]{Braylan2015}
Braylan A, Hollenbeck M, Meyerson E, Miikkulainen R (2015) {Reuse of Neural
  Modules for General Video Game Playing} \eprint{1512.01537}

\bibitem[{Bullmore and Sporns(2009)}]{Bullmore2009}
Bullmore E, Sporns O (2009) {Complex brain networks: graph theoretical analysis
  of structural and functional systems}. Nature Reviews Neuroscience
  10(3):186--198, \doi{10.1038/nrn2575}

\bibitem[{Bullmore and Bassett(2011)}]{Bullmore2011}
Bullmore ET, Bassett DS (2011) {Brain Graphs: Graphical Models of the Human
  Brain Connectome}. Annual Review of Clinical Psychology 7(1):113--140,
  \doi{10.1146/annurev-clinpsy-040510-143934}

\bibitem[{Buxhoeveden(2002)}]{Buxhoeveden2002}
Buxhoeveden DP (2002) {The minicolumn hypothesis in neuroscience}. Brain
  125(5):935--951, \doi{10.1093/brain/awf110}

\bibitem[{Caelli et~al.(1999)Caelli, Guan, and Wen}]{Caelli1999}
Caelli T, Guan L, Wen W (1999) {Modularity in neural computing}. Proceedings of
  the IEEE 87(9):1497--1518, \doi{10.1109/5.784227}

\bibitem[{Calabretta et~al.(2000)Calabretta, Nolfi, Parisi, and
  Wagner}]{Calabretta2000}
Calabretta R, Nolfi S, Parisi D, Wagner GP (2000) Duplication of modules
  facilitates the evolution of functional specialization. Artificial life
  6(1):69--84

\bibitem[{Chen et~al.(2008)Chen, He, Rosa-Neto, Germann, and Evans}]{Chen2008}
Chen ZJ, He Y, Rosa-Neto P, Germann J, Evans AC (2008) {Revealing modular
  architecture of human brain structural networks by using cortical thickness
  from MRI}. Cerebral Cortex 18(10):2374--2381, \doi{10.1093/cercor/bhn003}

\bibitem[{Chiang and Fu(1994)}]{Chiang1994}
Chiang CC, Fu HC (1994) A divide-and-conquer methodology for modular supervised
  neural network design. In: Neural Networks, 1994. IEEE World Congress on
  Computational Intelligence., 1994 IEEE International Conference on, IEEE,
  vol~1, pp 119--124

\bibitem[{Chihaoui et~al.(2016)Chihaoui, Elkefi, Bellil, and {Ben
  Amar}}]{Chihaoui2016}
Chihaoui M, Elkefi A, Bellil W, {Ben Amar} C (2016) {A Survey of 2D Face
  Recognition Techniques}. Computers 5(4):21, \doi{10.3390/computers5040021}

\bibitem[{Chollet(2016)}]{Chollet2016}
Chollet F (2016) {Xception: Deep Learning with Depthwise Separable
  Convolutions} \eprint{1610.02357}

\bibitem[{Ciregan et~al.(2012)Ciregan, Meier, and Schmidhuber}]{Ciregan2012}
Ciregan D, Meier U, Schmidhuber J (2012) Multi-column deep neural networks for
  image classification. In: Computer Vision and Pattern Recognition (CVPR),
  2012 IEEE Conference on, IEEE, pp 3642--3649

\bibitem[{Clune et~al.(2013)Clune, Mouret, and Lipson}]{Clune2013}
Clune J, Mouret JB, Lipson H (2013) {The evolutionary origins of modularity.}
  Proceedings Biological sciences / The Royal Society 280(1755):20122863,
  \doi{10.1098/rspb.2012.2863}, \eprint{1207.2743v1}

\bibitem[{Di~Ferdinando et~al.(2001)Di~Ferdinando, Calabretta, and
  Parisi}]{Di2001}
Di~Ferdinando A, Calabretta R, Parisi D (2001) Evolving modular architectures
  for neural networks

\bibitem[{Douglas and Martin(2007)}]{Douglas2007}
Douglas RJ, Martin KAC (2007) {Recurrent neuronal circuits in the neocortex.}
  Current biology : CB 17(13):R496--500, \doi{10.1016/j.cub.2007.04.024},
  \urlprefix\url{http://www.ncbi.nlm.nih.gov/pubmed/17610826}

\bibitem[{Eppel(2017)}]{Eppel2017}
Eppel S (2017) {Hierarchical semantic segmentation using modular convolutional
  neural networks} \urlprefix\url{https://arxiv.org/abs/1710.05126v1},
  \eprint{1710.05126}

\bibitem[{Eyben et~al.(2013)Eyben, Weninger, Squartini, and
  Schuller}]{Eyben2013}
Eyben F, Weninger F, Squartini S, Schuller B (2013) {Real-life voice activity
  detection with LSTM Recurrent Neural Networks and an application to Hollywood
  movies}. In: ICASSP, IEEE International Conference on Acoustics, Speech and
  Signal Processing - Proceedings, pp 483--487,
  \doi{10.1109/ICASSP.2013.6637694}

\bibitem[{Fernando et~al.(2017)Fernando, Banarse, Blundell, Zwols, Ha, Rusu,
  Pritzel, and Wierstra}]{Fernando2017}
Fernando C, Banarse D, Blundell C, Zwols Y, Ha D, Rusu AA, Pritzel A, Wierstra
  D (2017) {PathNet: Evolution Channels Gradient Descent in Super Neural
  Networks} \urlprefix\url{http://arxiv.org/abs/1701.08734},
  \eprint{1701.08734}

\bibitem[{Franco and Cannas(2001)}]{Franco2001}
Franco L, Cannas SA (2001) {Generalization properties of modular networks:
  Implementing the parity function}. IEEE Transactions on Neural Networks
  12(6):1306--1313, \doi{10.1109/72.963767}

\bibitem[{Freddolino et~al.(2008)Freddolino, Liu, Gruebele, and
  Schulten}]{Freddolino2008}
Freddolino PL, Liu F, Gruebele M, Schulten K (2008) {Ten-microsecond molecular
  dynamics simulation of a fast-folding WW domain.} Biophysical journal
  94(10):L75--7, \doi{10.1529/biophysj.108.131565},
  \urlprefix\url{http://www.ncbi.nlm.nih.gov/pubmed/18339748
  http://www.pubmedcentral.nih.gov/articlerender.fcgi?artid=PMC2367204}

\bibitem[{Fu et~al.(2001)Fu, Lee, Chiang, and Pao}]{Fu2001}
Fu HC, Lee YP, Chiang CC, Pao HT (2001) {Divide-and-conquer learning and
  modular perceptron networks}. IEEE Transactions on Neural Networks
  12(2):250--263, \doi{10.1109/72.914522}

\bibitem[{Garcia-Pedrajas et~al.(2003)Garcia-Pedrajas, Hervas-Martinez, and
  Munoz-Perez}]{Garcia-Pedrajas2003}
Garcia-Pedrajas N, Hervas-Martinez C, Munoz-Perez J (2003) {COVNET: a
  cooperative coevolutionary model for evolving artificial neural networks}.
  IEEE Transactions on Neural Networks 14(3):575--596,
  \doi{10.1109/TNN.2003.810618}

\bibitem[{Gollisch and Meister(2010)}]{Gollisch2010}
Gollisch T, Meister M (2010) {Eye Smarter than Scientists Believed: Neural
  Computations in Circuits of the Retina}. \doi{10.1016/j.neuron.2009.12.009},
  \eprint{NIHMS150003}

\bibitem[{Goltsev and Gritsenko(2015)}]{Goltsev2015}
Goltsev A, Gritsenko V (2015) {Modular neural networks with radial neural
  columnar architecture}. Biologically Inspired Cognitive Architectures
  13:63--74, \doi{10.1016/J.BICA.2015.06.001},
  \urlprefix\url{https://www.sciencedirect.com/science/article/pii/S2212683X15000286}

\bibitem[{Gradojevic et~al.(2009)Gradojevic, Gen{\c{c}}ay, and
  Kukolj}]{Gradojevic2009}
Gradojevic N, Gen{\c{c}}ay R, Kukolj D (2009) {Option pricing with modular
  neural networks.} IEEE transactions on neural networks / a publication of the
  IEEE Neural Networks Council 20(4):626--637, \doi{10.1109/TNN.2008.2011130}

\bibitem[{Guan and Li(2002)}]{Guan2002}
Guan SU, Li S (2002) Parallel growing and training of neural networks using
  output parallelism. IEEE transactions on Neural Networks 13(3):542--550

\bibitem[{{Guang-Bin Huang}(2003)}]{Guang-BinHuang2003}
{Guang-Bin Huang} (2003) {Learning capability and storage capacity of
  two-hidden-layer feedforward networks}. IEEE Transactions on Neural Networks
  14(2):274--281, \doi{10.1109/TNN.2003.809401}

\bibitem[{Happel and Murre(1994)}]{Happel1994}
Happel BLM, Murre JMJ (1994) {Design and evolution of modular neural network
  architectures}. Neural Networks 7(6-7):985--1004,
  \doi{10.1016/S0893-6080(05)80155-8}

\bibitem[{He et~al.(2016)He, Zhang, Ren, and Sun}]{He2016}
He K, Zhang X, Ren S, Sun J (2016) Deep residual learning for image
  recognition. In: Proceedings of the IEEE conference on computer vision and
  pattern recognition, pp 770--778

\bibitem[{Hidalgo et~al.(2009)Hidalgo, Castillo, and Melin}]{Hidalgo2009}
Hidalgo D, Castillo O, Melin P (2009) Type-1 and type-2 fuzzy inference systems
  as integration methods in modular neural networks for multimodal biometry and
  its optimization with genetic algorithms. Information Sciences
  179(13):2123--2145

\bibitem[{Hochreiter and {Urgen Schmidhuber}(1997)}]{Hochreiter1997}
Hochreiter S, {Urgen Schmidhuber} J (1997) {LONG SHORT-TERM MEMORY}. Neural
  Computation 9(8):1735--1780, \doi{10.1162/neco.1997.9.8.1735},
  \eprint{1206.2944}

\bibitem[{Hu et~al.(2016)Hu, Rohrbach, Andreas, Darrell, and Saenko}]{Hu2016}
Hu R, Rohrbach M, Andreas J, Darrell T, Saenko K (2016) {Modeling Relationships
  in Referential Expressions with Compositional Modular Networks}
  \eprint{1611.09978}

\bibitem[{Huang et~al.(2016)Huang, Sun, Liu, Sedra, and Weinberger}]{Huang2016}
Huang G, Sun Y, Liu Z, Sedra D, Weinberger KQ (2016) Deep networks with
  stochastic depth. In: European Conference on Computer Vision, Springer, pp
  646--661

\bibitem[{Huizinga et~al.(2014)Huizinga, Mouret, and Clune}]{Huizinga2014}
Huizinga J, Mouret JB, Clune J (2014) {Evolving Neural Networks That Are Both
  Modular and Regular: HyperNeat Plus the Connection Cost Technique}. Gecco pp
  697--704, \doi{10.1145/2576768.2598232}

\bibitem[{H{\"u}sken et~al.(2002)H{\"u}sken, Igel, and Toussaint}]{Husken2002}
H{\"u}sken M, Igel C, Toussaint M (2002) Task-dependent evolution of modularity
  in neural networks. Connection Science 14(3):219--229

\bibitem[{Kaiser and Hilgetag(2010)}]{Kaiser2010}
Kaiser M, Hilgetag CC (2010) Optimal hierarchical modular topologies for
  producing limited sustained activation of neural networks. Frontiers in
  neuroinformatics 4

\bibitem[{Karami et~al.(2013)Karami, Safabakhsh, and Rahmati}]{Karami2013}
Karami M, Safabakhsh R, Rahmati M (2013) {Modular cellular neural network
  structure for wave-computing-based image processing}. ETRI Journal
  35(2):207--217, \doi{10.4218/etrij.13.0112.0107}

\bibitem[{Kashtan and Alon(2005)}]{Kashtan2005}
Kashtan N, Alon U (2005) {Spontaneous evolution of modularity and network
  motifs.} Proceedings of the National Academy of Sciences of the United States
  of America 102(39):13773--8, \doi{10.1073/pnas.0503610102}

\bibitem[{Kastellakis et~al.(2015)Kastellakis, Cai, Mednick, Silva, and
  Poirazi}]{Kastellakis2015}
Kastellakis G, Cai DJ, Mednick SC, Silva AJ, Poirazi P (2015) {Synaptic
  clustering within dendrites: An emerging theory of memory formation}.
  \doi{10.1016/j.pneurobio.2014.12.002}, \eprint{15334406}

\bibitem[{{Kien Tuong Phan} et~al.(2015){Kien Tuong Phan}, Maul, and {Tuong
  Thuy Vu}}]{KienTuongPhan2015}
{Kien Tuong Phan}, Maul TH, {Tuong Thuy Vu} (2015) {A parallel circuit approach
  for improving the speed and generalization properties of neural networks}.
  In: 2015 11th International Conference on Natural Computation (ICNC), IEEE,
  pp 1--7, \doi{10.1109/ICNC.2015.7377956}

\bibitem[{Kim et~al.(2017)Kim, Cha, Kim, Lee, and Kim}]{Kim2017}
Kim T, Cha M, Kim H, Lee JK, Kim J (2017) {Learning to Discover Cross-Domain
  Relations with Generative Adversarial Networks} \eprint{1703.05192}

\bibitem[{Larsson et~al.(2016)Larsson, Maire, and Shakhnarovich}]{Larsson2016}
Larsson G, Maire M, Shakhnarovich G (2016) {FractalNet: Ultra-Deep Neural
  Networks without Residuals} \eprint{1605.07648}

\bibitem[{Lin et~al.(2013)Lin, Chen, and Yan}]{Lin2013}
Lin M, Chen Q, Yan S (2013) {Network In Network}. arXiv preprint p~10,
  \doi{10.1109/ASRU.2015.7404828}, \eprint{1312.4400}

\bibitem[{Lodato and Arlotta(2015)}]{Lodato2015}
Lodato S, Arlotta P (2015) {Generating Neuronal Diversity in the Mammalian
  Cerebral Cortex}. Annual Review of Cell and Developmental Biology
  31(1):699--720, \doi{10.1146/annurev-cellbio-100814-125353},
  \eprint{15334406}

\bibitem[{L{\'{o}}pez-Mu{\~{n}}oz et~al.(2006)L{\'{o}}pez-Mu{\~{n}}oz, Boya,
  and Alamo}]{Lopez-Munoz2006}
L{\'{o}}pez-Mu{\~{n}}oz F, Boya J, Alamo C (2006) {Neuron theory, the
  cornerstone of neuroscience, on the centenary of the Nobel Prize award to
  Santiago Ram{\'{o}}n y Cajal}. Brain Research Bulletin 70(4-6):391--405,
  \doi{10.1016/j.brainresbull.2006.07.010}

\bibitem[{Melin et~al.(2007)Melin, Mancilla, Lopez, and Mendoza}]{Melin2007}
Melin P, Mancilla A, Lopez M, Mendoza O (2007) A hybrid modular neural network
  architecture with fuzzy sugeno integration for time series forecasting.
  Applied Soft Computing 7(4):1217--1226

\bibitem[{Melin et~al.(2011)Melin, Mendoza, and Castillo}]{Melin2011}
Melin P, Mendoza O, Castillo O (2011) Face recognition with an improved
  interval type-2 fuzzy logic sugeno integral and modular neural networks. IEEE
  Transactions on systems, man, and cybernetics-Part A: systems and humans
  41(5):1001--1012

\bibitem[{Mendoza et~al.(2009a)Mendoza, Melin, and Licea}]{Mendoza2009a}
Mendoza O, Melin P, Licea G (2009a) {A hybrid approach for image recognition
  combining type-2 fuzzy logic, modular neural networks and the Sugeno
  integral}. Information Sciences 179(13):2078--2101,
  \doi{10.1016/j.ins.2008.11.018}

\bibitem[{Mendoza et~al.(2009b)Mendoza, Mel{\'{i}}n, and
  Castillo}]{Mendoza2009b}
Mendoza O, Mel{\'{i}}n P, Castillo O (2009b) {Interval type-2 fuzzy logic and
  modular neural networks for face recognition applications}. Applied Soft
  Computing 9(4):1377--1387, \doi{10.1016/j.asoc.2009.06.007}

\bibitem[{Meunier et~al.(2010)Meunier, Lambiotte, and Bullmore}]{Meunier2010}
Meunier D, Lambiotte R, Bullmore ET (2010) {Modular and hierarchically modular
  organization of brain networks}. \doi{10.3389/fnins.2010.00200}

\bibitem[{Miikkulainen et~al.(2017)Miikkulainen, Liang, Meyerson, Rawal, Fink,
  Francon, Raju, Shahrzad, Navruzyan, Duffy, and Hodjat}]{Miikkulainen2017}
Miikkulainen R, Liang J, Meyerson E, Rawal A, Fink D, Francon O, Raju B,
  Shahrzad H, Navruzyan A, Duffy N, Hodjat B (2017) {Evolving Deep Neural
  Networks} \eprint{1703.00548}

\bibitem[{Mountcastle(1997)}]{Mountcastle1997}
Mountcastle VB (1997) {The columnar organization of the neocortex}.
  \doi{10.1093/brain/120.4.701}

\bibitem[{Mouret and Doncieux(2008)}]{Mouret2008}
Mouret JB, Doncieux S (2008) {MENNAG: a modular, regular and hierarchical
  encoding for neural-networks based on attribute grammars}. Evolutionary
  Intelligence 1(3):187--207, \doi{10.1007/s12065-008-0015-7}

\bibitem[{Mouret and Doncieux(2009)}]{Mouret2009}
Mouret JB, Doncieux S (2009) {Evolving modular neural-networks through
  exaptation}. In: 2009 IEEE Congress on Evolutionary Computation, CEC 2009, pp
  1570--1577, \doi{10.1109/CEC.2009.4983129}

\bibitem[{de~Nardi et~al.(2006)de~Nardi, Togelius, Holland, and
  Lucas}]{DeNardi}
de~Nardi R, Togelius J, Holland O, Lucas S (2006) {Evolution of Neural Networks
  for Helicopter Control: Why Modularity Matters}. In: 2006 IEEE International
  Conference on Evolutionary Computation, IEEE, pp 1799--1806,
  \doi{10.1109/CEC.2006.1688525}

\bibitem[{Newman(2004)}]{Newman2004}
Newman MEJ (2004) {Detecting community structure in networks}. Eur Phys J B
  38:321--330, \doi{10.1140/epjb/e2004-00124-y}

\bibitem[{Newman(2006)}]{Newman2006}
Newman MEJ (2006) {Modularity and community structure in networks.} Proceedings
  of the National Academy of Sciences of the United States of America
  103(23):8577--82, \doi{10.1073/pnas.0601602103}

\bibitem[{Newman(2016)}]{Newman2016}
Newman MEJ (2016) {Community detection in networks: Modularity optimization and
  maximum likelihood are equivalent}. Arvix 1:1--8,
  \doi{10.1103/PhysRevE.94.052315}, \eprint{1606.02319}

\bibitem[{Oh and Suen(2002)}]{Oh2002}
Oh IS, Suen CY (2002) {A class-modular feedforward neural network for
  handwriting recognition}. Pattern Recognition 35(1):229--244,
  \doi{10.1016/S0031-3203(00)00181-3}

\bibitem[{Ort{\'{i}}n et~al.(2005)Ort{\'{i}}n, Guti{\'{e}}rrez, Pesquera, and
  Vasquez}]{Ortin2005}
Ort{\'{i}}n S, Guti{\'{e}}rrez J, Pesquera L, Vasquez H (2005) {Nonlinear
  dynamics extraction for time-delay systems using modular neural networks
  synchronization and prediction}. Physica A: Statistical Mechanics and its
  Applications 351(1):133--141, \doi{10.1016/j.physa.2004.12.015}

\bibitem[{Ou and Murphey(2007)}]{Ou2007}
Ou G, Murphey YL (2007) {Multi-class pattern classification using neural
  networks}. Pattern Recognition 40(1):4--18,
  \doi{10.1016/j.patcog.2006.04.041}

\bibitem[{Pan et~al.(2016)Pan, Xu, Yang, Wu, and Zhuang}]{Pan2016}
Pan P, Xu Z, Yang Y, Wu F, Zhuang Y (2016) {Hierarchical Recurrent Neural
  Encoder for Video Representation With Application to Captioning}. In: The
  IEEE Conference on Computer Vision and Pattern Recognition (CVPR)

\bibitem[{Phan et~al.(2016)Phan, Maul, Vu, and Lai}]{Phan2016}
Phan KT, Maul TH, Vu TT, Lai WK (2016) {Improving neural network generalization
  by combining parallel circuits with dropout}. In: Lecture Notes in Computer
  Science (including subseries Lecture Notes in Artificial Intelligence and
  Lecture Notes in Bioinformatics), vol 9949 LNCS, pp 572--580,
  \doi{10.1007/978-3-319-46675-0_63}, \eprint{1612.04970}

\bibitem[{Phan et~al.(2017)Phan, Maul, Vu, and Lai}]{Phan2017}
Phan KT, Maul TH, Vu TT, Lai WK (2017) Dropcircuit: A modular regularizer for
  parallel circuit networks. Neural Processing Letters pp 1--18

\bibitem[{{Phyo Phyo San} et~al.(2011){Phyo Phyo San}, {Sai Ho Ling}, and
  Nguyen}]{PhyoPhyoSan2011}
{Phyo Phyo San}, {Sai Ho Ling}, Nguyen HT (2011) {Block based neural network
  for hypoglycemia detection}. In: 2011 Annual International Conference of the
  IEEE Engineering in Medicine and Biology Society, IEEE, pp 5666--5669,
  \doi{10.1109/IEMBS.2011.6091371}

\bibitem[{Radicchi et~al.(2004)Radicchi, Castellano, Cecconi, Loreto, and
  Parisi}]{Radicchi2003}
Radicchi F, Castellano C, Cecconi F, Loreto V, Parisi D (2004) Defining and
  identifying communities in networks. Proceedings of the National Academy of
  Sciences of the United States of America 101(9):2658--2663

\bibitem[{Reisinger et~al.(2004)Reisinger, Stanley, and
  Miikkulainen}]{Reisinger2004}
Reisinger J, Stanley KO, Miikkulainen R (2004) Evolving reusable neural
  modules. In: Genetic and Evolutionary Computation Conference, Springer, pp
  69--81

\bibitem[{Ronen et~al.(2002)Ronen, Shabtai, and Guterman}]{Ronen2002}
Ronen M, Shabtai Y, Guterman H (2002) {Hybrid model building methodology using
  unsupervised fuzzy clustering and supervised neural networks}. Biotechnol
  Bioeng 77(4):420--429

\bibitem[{Rudasi and Zahorian(1991)}]{Rudasi1991}
Rudasi L, Zahorian S (1991) {Text-independent talker identification with neural
  networks}. In: [Proceedings] ICASSP 91: 1991 International Conference on
  Acoustics, Speech, and Signal Processing, IEEE, pp 389--392 vol.1,
  \doi{10.1109/ICASSP.1991.150358}

\bibitem[{Sabour et~al.(2017)Sabour, Frosst, and Hinton}]{Sabour2017}
Sabour S, Frosst N, Hinton GE (2017) {Dynamic Routing Between Capsules}.
  \urlprefix\url{http://papers.nips.cc/paper/6975-dynamic-routing-between-capsules}

\bibitem[{{Sang-Woo Moon} and {Seong-Gon Kong}(2001)}]{Sang-WooMoon2001}
{Sang-Woo Moon}, {Seong-Gon Kong} (2001) {Block-based neural networks}. IEEE
  Transactions on Neural Networks 12(2):307--317, \doi{10.1109/72.914525}

\bibitem[{Schwarz et~al.(2008)Schwarz, Gozzi, and Bifone}]{Schwarz2008}
Schwarz AJ, Gozzi A, Bifone A (2008) {Community structure and modularity in
  networks of correlated brain activity}. Magnetic Resonance Imaging
  26(7):914--920, \doi{10.1016/j.mri.2008.01.048}, \eprint{0701041v2}

\bibitem[{Serban et~al.(2016)Serban, Sordoni, Bengio, Courville, and
  Pineau}]{Serban2016}
Serban IV, Sordoni A, Bengio Y, Courville A, Pineau J (2016) {Building
  End-To-End Dialogue Systems Using Generative Hierarchical Neural Network
  Models}. Aaai p~8, \doi{10.1017/CBO9781107415324.004}, \eprint{1507.04808}

\bibitem[{Sharkey(1996)}]{Sharkey1996}
Sharkey AJC (1996) {On Combining Artificial Neural Nets}. Connection Science
  8(3-4):299--313, \doi{10.1080/095400996116785}

\bibitem[{Shetty and Laaksonen(2015)}]{Shetty2015}
Shetty R, Laaksonen J (2015) {Video captioning with recurrent networks based on
  frame- and video-level features and visual content classification}
  \eprint{1512.02949}

\bibitem[{Singh et~al.(2016)Singh, Hoiem, and Forsyth}]{Singh2016}
Singh S, Hoiem D, Forsyth D (2016) Swapout: Learning an ensemble of deep
  architectures. In: Advances in Neural Information Processing Systems, pp
  28--36

\bibitem[{Soutner and M{\"u}ller(2013)}]{Soutner2013}
Soutner D, M{\"u}ller L (2013) Application of lstm neural networks in language
  modelling. In: International Conference on Text, Speech and Dialogue,
  Springer, pp 105--112

\bibitem[{Sporns(2011)}]{Sporns2011}
Sporns O (2011) {The human connectome: A complex network}.
  \doi{10.1111/j.1749-6632.2010.05888.x}

\bibitem[{Sporns and Zwi(2004)}]{Sporns2004}
Sporns O, Zwi JD (2004) {The Small World of the Cerebral Cortex}.
  Neuroinformatics 2(2):145--162, \doi{10.1385/NI:2:2:145}

\bibitem[{Srivastava et~al.(2014)Srivastava, Hinton, Krizhevsky, Sutskever, and
  Salakhutdinov}]{Srivastava2014}
Srivastava N, Hinton G, Krizhevsky A, Sutskever I, Salakhutdinov R (2014)
  {Dropout: A Simple Way to Prevent Neural Networks from Overfitting}. Journal
  of Machine Learning Research 15:1929--1958, \doi{10.1214/12-AOS1000},
  \eprint{1102.4807}

\bibitem[{Srivastava et~al.(2013)Srivastava, Masci, Kazerounian, Gomez, and
  Schmidhuber}]{Srivastava2013}
Srivastava RK, Masci J, Kazerounian S, Gomez F, Schmidhuber J (2013) {Compete
  to Compute}. Nips pp 2310--2318

\bibitem[{Srivastava et~al.(2015)Srivastava, Greff, and
  Schmidhuber}]{Srivastava2015}
Srivastava RK, Greff K, Schmidhuber J (2015) {Highway Networks}.
  arXiv:150500387 [cs] \eprint{1505.00387}

\bibitem[{Stanley and Miikkulainen(2002)}]{Stanley2002}
Stanley KO, Miikkulainen R (2002) {Evolving Neural Networks through Augmenting
  Topologies}. Evolutionary Computation 10(2):99--127,
  \doi{10.1162/106365602320169811}, \eprint{1407.0576}

\bibitem[{Stanley et~al.(2009)Stanley, D'Ambrosio, and Gauci}]{Stanley2009}
Stanley KO, D'Ambrosio DB, Gauci J (2009) {A Hypercube-Based Encoding for
  Evolving Large-Scale Neural Networks}. Artificial Life 15(2):185--212,
  \doi{10.1162/artl.2009.15.2.15202}

\bibitem[{Stollenga et~al.(2015)Stollenga, Byeon, Liwicki, and
  Schmidhuber}]{Stollenga2015}
Stollenga MF, Byeon W, Liwicki M, Schmidhuber J (2015) {Parallel
  Multi-Dimensional LSTM, With Application to Fast Biomedical Volumetric Image
  Segmentation}. In: Cortes C, Lawrence ND, Lee DD, Sugiyama M, Garnett R (eds)
  Advances in Neural Information Processing Systems 28, Curran Associates,
  Inc., pp 2998--3006

\bibitem[{Subirats et~al.(2010)Subirats, Jerez, G{\'{o}}mez, and
  Franco}]{Subirats2010}
Subirats JL, Jerez JM, G{\'{o}}mez I, Franco L (2010) {Multiclass Pattern
  Recognition Extension for the New C-Mantec Constructive Neural Network
  Algorithm}. Cognitive Computation 2(4):285--290,
  \doi{10.1007/s12559-010-9051-6}

\bibitem[{Szegedy et~al.(2015)Szegedy, Liu, Jia, Sermanet, Reed, Anguelov,
  Erhan, Vanhoucke, and Rabinovich}]{Szegedy2015}
Szegedy C, Liu W, Jia Y, Sermanet P, Reed S, Anguelov D, Erhan D, Vanhoucke V,
  Rabinovich A (2015) {Going deeper with convolutions}. In: Proceedings of the
  IEEE Computer Society Conference on Computer Vision and Pattern Recognition,
  vol 07-12-June, pp 1--9, \doi{10.1109/CVPR.2015.7298594}, \eprint{1409.4842}

\bibitem[{Szegedy et~al.(2016)Szegedy, Vanhoucke, Ioffe, Shlens, and
  Wojna}]{Szegedy2016}
Szegedy C, Vanhoucke V, Ioffe S, Shlens J, Wojna Z (2016) Rethinking the
  inception architecture for computer vision. In: Proceedings of the IEEE
  Conference on Computer Vision and Pattern Recognition, pp 2818--2826

\bibitem[{Terekhov et~al.(2015)Terekhov, Montone, and O'Regan}]{Terekhov2015}
Terekhov AV, Montone G, O'Regan JK (2015) {Knowledge Transfer in Deep
  Block-Modular Neural Networks}. Springer, Cham, pp 268--279,
  \doi{10.1007/978-3-319-22979-9_27},
  \urlprefix\url{http://link.springer.com/10.1007/978-3-319-22979-9{\_}27}

\bibitem[{Tyler et~al.(2005)Tyler, Wilkinson, and Huberman}]{Tyler2005}
Tyler JR, Wilkinson DM, Huberman BA (2005) {E-Mail as Spectroscopy: Automated
  Discovery of Community Structure within Organizations}. The Information
  Society 21(2):143--153, \doi{10.1080/01972240590925348}, \eprint{0303264}

\bibitem[{Veit et~al.(2016)Veit, Wilber, and Belongie}]{Veit2016}
Veit A, Wilber MJ, Belongie S (2016) Residual networks behave like ensembles of
  relatively shallow networks. In: Advances in Neural Information Processing
  Systems, pp 550--558

\bibitem[{Verbancsics and Stanley(2011)}]{Verbancsics2011}
Verbancsics P, Stanley KO (2011) {Constraining connectivity to encourage
  modularity in HyperNEAT}. Proceedings of the 13th annual conference on
  Genetic and evolutionary computation - GECCO '11 p 1483,
  \doi{10.1145/2001576.2001776}

\bibitem[{Vlahogianni et~al.(2007)Vlahogianni, Karlaftis, and
  Golias}]{Vlahogianni2007}
Vlahogianni EI, Karlaftis MG, Golias JC (2007) Spatio-temporal short-term urban
  traffic volume forecasting using genetically optimized modular networks.
  Computer-Aided Civil and Infrastructure Engineering 22(5):317--325

\bibitem[{Wang(2015)}]{Wang2015}
Wang M (2015) {Multi-path Convolutional Neural Networks for Complex Image
  Classification} \eprint{1506.04701}

\bibitem[{Wang et~al.(2011)Wang, Hilgetag, and Zhou}]{Wang2011}
Wang SJ, Hilgetag CC, Zhou C (2011) Sustained activity in hierarchical modular
  neural networks: self-organized criticality and oscillations. Frontiers in
  computational neuroscience 5

\bibitem[{Wang et~al.(2012)Wang, Wu, Coates, and Ng}]{Wang2012}
Wang T, Wu DJ, Coates A, Ng AY (2012) End-to-end text recognition with
  convolutional neural networks. In: Pattern Recognition (ICPR), 2012 21st
  International Conference on, IEEE, pp 3304--3308

\bibitem[{Watts(1999)}]{Watts1999}
Watts DJ (1999) {Networks, Dynamics, and the Small‐World Phenomenon}.
  American Journal of Sociology 105(2):493--527, \doi{10.1086/210318},
  \eprint{9910332}

\bibitem[{{Wei Jiang} and {Seong Kong}(2007)}]{WeiJiang2007}
{Wei Jiang}, {Seong Kong} G (2007) {Block-Based Neural Networks for
  Personalized ECG Signal Classification}. IEEE Transactions on Neural Networks
  18(6):1750--1761, \doi{10.1109/TNN.2007.900239}

\bibitem[{Weston et~al.(2014)Weston, Chopra, and Bordes}]{Weston2014}
Weston J, Chopra S, Bordes A (2014) {Memory Networks} \eprint{1410.3916}

\bibitem[{Xie et~al.(2016)Xie, Girshick, Doll{\'a}r, Tu, and He}]{Xie2016}
Xie S, Girshick R, Doll{\'a}r P, Tu Z, He K (2016) Aggregated residual
  transformations for deep neural networks. arXiv preprint arXiv:161105431

\bibitem[{Xu et~al.(1992)Xu, Krzyzak, and Suen}]{Xu1992}
Xu L, Krzyzak A, Suen C (1992) {Methods of combining multiple classifiers and
  their applications to handwriting recognition}. IEEE Transactions on Systems,
  Man, and Cybernetics 22(3):418--435, \doi{10.1109/21.155943}

\bibitem[{Yu et~al.(2016)Yu, Wang, Huang, Yang, and Xu}]{Yu2016}
Yu H, Wang J, Huang Z, Yang Y, Xu W (2016) {Video Paragraph Captioning Using
  Hierarchical Recurrent Neural Networks}. In: The IEEE Conference on Computer
  Vision and Pattern Recognition (CVPR)

\bibitem[{Yu et~al.(2018)Yu, Lin, Shen, Yang, Lu, Bansal, and Berg}]{Yu2018}
Yu L, Lin Z, Shen X, Yang J, Lu X, Bansal M, Berg TL (2018) {MAttNet: Modular
  Attention Network for Referring Expression Comprehension}
  \urlprefix\url{https://arxiv.org/abs/1801.08186v2}, \eprint{1801.08186}

\bibitem[{Zhang et~al.(2016)Zhang, Leitner, Milford, and Corke}]{Zhang2016}
Zhang F, Leitner J, Milford M, Corke P (2016) {Modular Deep Q Networks for
  Sim-to-real Transfer of Visuo-motor Policies}
  \urlprefix\url{https://arxiv.org/abs/1610.06781v4}, \eprint{1610.06781}

\bibitem[{Zhang et~al.(2014)Zhang, Donahue, Girshick, and Darrell}]{Zhang2014}
Zhang N, Donahue J, Girshick R, Darrell T (2014) {Part-based R-CNNs for
  fine-grained category detection}. In: Lecture Notes in Computer Science
  (including subseries Lecture Notes in Artificial Intelligence and Lecture
  Notes in Bioinformatics), vol 8689 LNCS, pp 834--849,
  \doi{10.1007/978-3-319-10590-1_54}, \eprint{1407.3867}

\bibitem[{Zheng et~al.(2006)Zheng, Lee, and Shi}]{Zheng2006}
Zheng W, Lee DH, Shi Q (2006) {Short-Term Freeway Traffic Flow Prediction:
  Bayesian Combined Neural Network Approach}. Journal of Transportation
  Engineering 132(2):114--121, \doi{10.1061/(ASCE)0733-947X(2006)132:2(114)}

\end{thebibliography}

% \begin{thebibliography}{}
% %
% % and use \bibitem to create references. Consult the Instructions
% % for authors for reference list style.
% % Format for Journal Reference
% Author, Article title, Journal, Volume, page numbers (year)
% Author, Book title, page numbers. Publisher, place (year)
% etc
%\end{thebibliography}

\end{document}